\pdfoutput=1
\documentclass[journal]{IEEEtran}
\usepackage{amsmath,amsfonts,amssymb}
\usepackage{algorithmic}
\usepackage{algorithm}
\usepackage{array}
\usepackage[caption=false,font=normalsize,labelfont=sf,textfont=sf]{subfig}
\usepackage{textcomp}
\usepackage{stfloats}
\usepackage{url}
\usepackage{verbatim}
\usepackage{graphicx}
\usepackage{booktabs}
\usepackage{multirow}
\usepackage[table]{xcolor}
\usepackage{cite}
\definecolor{priCol}{RGB}{240,245,252}
\definecolor{oursRow}{RGB}{232,242,254}
\definecolor{gainGreen}{RGB}{0,120,50}
\newcommand{\gain}[1]{\,{\tiny\textcolor{gainGreen}{\textbf{+#1}}}}
\begin{document}

\title{JevNavAgent: A Single-Forward Visual Navigation Agent with Trajectory-Level RL for Gigapixel Pathology}

\author{Yonghuang Wu, Zhenyang Liang, Pengfei Wang, Yize Ma, Guoqing Wu, and Jinhua Yu%
\thanks{Yonghuang Wu, Zhenyang Liang, Pengfei Wang, Yize Ma, Guoqing Wu, and Jinhua Yu are with the School of Biomedical Engineering and Technological Innovation, Fudan University, Shanghai, China (Corresponding author: Jinhua Yu, e-mail: jhyu@fudan.edu.cn).}}

\markboth{IEEE Transactions Draft}%
{JevNavAgent: A Single-Forward Visual Navigation Agent with Trajectory-Level RL for Gigapixel Pathology}

\maketitle

\begin{abstract}
Gigapixel whole-slide images (WSIs) exceed single-pass vision-language model (VLM) capacity at cellular resolution, while multiple instance learning (MIL) relies on exhaustive, query-agnostic offline patching without interactive cross-scale reasoning. Autonomous multi-step VLM navigation agents inspect diagnostic regions on demand, yet face two core bottlenecks: (1) multi-turn autoregressive action generation incurs heavy per-step decoding latency, throttling exploration and making online multi-trajectory agentic reinforcement learning (RL) costly; and (2) offline trajectory distillation suffers from an existential-versus-universal stopping asymmetry between positive detection and negative exclusion, inducing thumbnail and kinematic shortcuts while leaving imitation policies prone to passing through metastases without stopping. In the Jev (System One) paradigm, a model predicts distributions over predefined decision spaces in one forward pass per state (0 generated tokens) while deterministic code executes spatial navigation. On this principle, we build JevNavAgent, a 2B-parameter agent combining factorized stopping, spatial-candidate, and contextual-bilinear answer heads with overview modality dropout. We pair Stage-1 offline counterfactual trajectory alignment (mitigating depth and panning shortcuts) with Stage-2 online trajectory-level GRPO (Traj-GRPO), using decoding-free rollouts to optimize branch-level lesion discovery, in-polygon stopping, and step-wise confidence gain for VQA and screening. On Clean Visual WSI-VQA, JevNavAgent reaches 51.57\% multi-scale (50.43\% terminal-step) overall and 51.39\% focal accuracy, outperforming frontier VLM agents and 4B--11B open-source cascades while running nearly 12\texttimes\ faster than frontier API agents (5.63\,s vs.\ 66.64--70.04\,s). On CAMELYON16, Stage-2 Traj-GRPO reaches 22.45\% Official Test localization (converting 11/12, or 91.7\%, of visited tumors into grounded stops) with 93.75\% specificity and 69.77\% slide accuracy, outperforming Stage-1 SFT under both matched stopping rules alongside zero-shot budget scaling.
\end{abstract}

\begin{IEEEkeywords}
Computational pathology, whole-slide imaging, vision-language models, autonomous navigation agent, agentic reinforcement learning, counterfactual trajectory alignment.
\end{IEEEkeywords}

\section{Introduction}
\label{sec:intro}

\IEEEPARstart{W}{hole-slide} images (WSIs) in digital pathology routinely exceed $100{,}000 \times 100{,}000$ pixels at $40\times$ magnification ($0.25\,\mu\text{m/pixel}$)~\cite{bejnordi2017diagnostic,campanella2019clinical}. A single resized whole-slide overview thumbnail (or an initial $512 \times 512$ crop at $\text{DS}=64$) preserves tissue layout but obliterates cellular pleomorphism, mitotic figures, and sub-millimeter lymph node micrometastases~\cite{chen2022scaling,lu2021data}. Conversely, tiling a slide into tens of thousands of high-magnification patches for multiple instance learning (MIL)~\cite{ilse2018attention,shao2021transmil} or whole-slide foundation encoders~\cite{xu2024whole,chen2024uni} requires costly offline feature extraction that is agnostic to the clinical query, leaving no room for interactive, query-conditioned cross-scale inspection.

\begin{figure*}[t]
\centering
\includegraphics[width=0.98\textwidth]{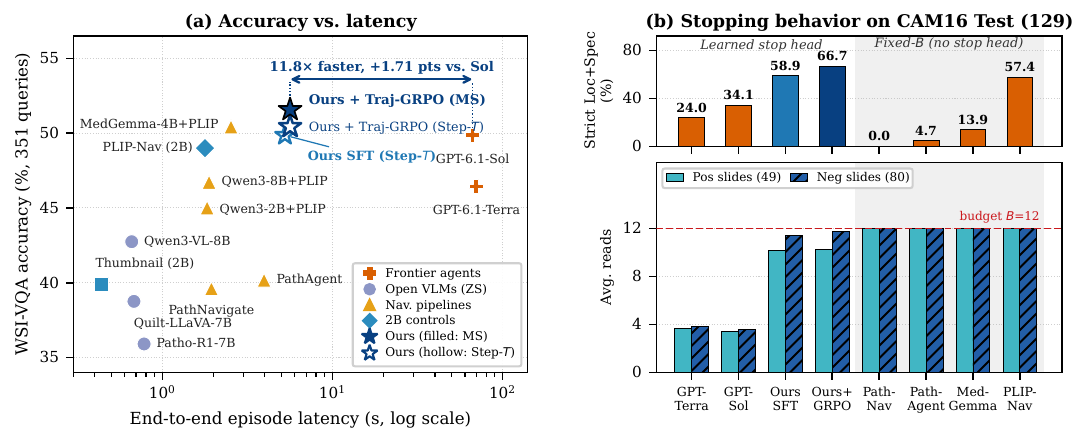}
\caption{Accuracy--efficiency trade-off and stopping behavior of gigapixel pathology navigation paradigms. \textbf{(a)} Clean Visual WSI-VQA ($351$ questions): JevNavAgent + Traj-GRPO attains the highest multi-scale accuracy ($51.57\%$, filled star) and focal accuracy ($51.39\%$), with Step-$T$ accuracy ($50.43\%$, hollow star) on par with the strongest 4B baseline, at $5.63\,\text{s}$ per episode, $11.8\times\text{--}12.4\times$ faster than frontier multi-turn agents (GPT-6.1-Sol/Terra); the SFT variant is shown with Step-$T$ readout. \textbf{(b)} CAMELYON16 Official Test ($129$ WSIs, $B=12$): frontier agents stop after $\sim 3.6$ reads on both positive and negative slides, and fixed-$B$ pipelines (shaded) have no stopping head and always exhaust the budget (dashed line), whereas JevNavAgent + Traj-GRPO stops asymmetrically ($10.29$ vs.\ $11.75$ reads on positive vs.\ negative slides). Top bars: Strict Loc+Spec ($\%$); bottom bars: mean reads on positive (solid) and negative (hatched) slides.}
\label{fig:intro_pareto}
\end{figure*}

Human pathologists work differently: they scan a low-power overview ($0.625\times\text{--}2.5\times$), zoom into suspicious regions ($10\times\text{--}40\times$), pan along tissue margins, and backtrack when a candidate proves benign. This motivates formulating gigapixel WSI interpretation as an autonomous multi-step visual navigation agent~\cite{mnih2014recurrent,ba2014multiple}. Yet our systematic evaluation on $351$ held-out WSI-VQA queries and CAMELYON16 ($46$ Dev and $129$ Official Test slides from $311$ verified slides; Fig.~\ref{fig:intro_pareto}) shows that both zero-shot frontier vision-language model (VLM) agents and standard supervised fine-tuning (SFT) of compact open-source VLMs struggle to navigate effectively, owing to two interlocking challenges:
\begin{enumerate}
    \item \textbf{Decoding Overhead and Agentic RL Scalability:} Multi-turn VLM agents serialize coordinates and reasoning into dozens of autoregressive tokens per step. The accumulated latency ($66.64\text{--}70.04\,\text{s}$ per slide for the frontier agents in Fig.~\ref{fig:intro_pareto}a) throttles interactive exploration and makes online multi-trajectory agentic reinforcement learning (RL) computationally costly. Conversely, scoring answer options solely from static token embeddings avoids decoding, but sacrifices contextual language representations.
    \item \textbf{Stopping Asymmetry, Distillation Shortcuts, and Trajectory-Level Stopping Gap:} Clinical screening is inherently asymmetric: a positive call is an existential proof (one verified lesion crop suffices), whereas a negative call is a universal exclusion proof (multiple suspicious regions must be ruled out). Frontier agents halt prematurely after $3.5\text{--}4.4$ reads on a single tissue root ($27.5\%\text{--}52.2\%$ negative specificity on CAMELYON16), and multi-scale cascades~\cite{huang2023visual} have no stopping mechanism under fixed-budget evaluation (``Fixed-$B$'' in Fig.~\ref{fig:intro_pareto}b). Distilling such trajectories into a compact student further leaks non-visual cues~\cite{ross2011reduction,geirhos2020shortcut}, letting the student exploit search depth, terminal scale transitions, or the overview thumbnail instead of cellular evidence. Even after state-level debiasing, offline behavioral cloning lacks trajectory-level credit assignment and exhibits an in-polygon stopping gap on CAMELYON16 Test: it visits a true metastasis along a branch (``Visited Hit'') yet fails to stop before drifting back into benign tissue.
\end{enumerate}

To overcome these bottlenecks, we draw on the Jev (System One) paradigm, in which a model maps a multimodal context directly to probability distributions over predefined decision spaces in a single forward pass ($0$ generated tokens), while deterministic code orchestrates the surrounding workflow. Building on it, we develop JevNavAgent, an autonomous visual navigation agent on a compact 2B-parameter VLM backbone (Qwen3-VL-2B) that outputs stopping, candidate-view, and answer distributions in one pass, with deterministic code executing the multi-scale navigation. In the spirit of the Jevons paradox, removing autoregressive decoding lowers per-step cost enough to make two things practical: thorough multi-region verification at inference time, and online multi-trajectory rollouts for agentic RL at training time. Our contributions are as follows:
\begin{enumerate}
    \item \textbf{Single-Forward Navigation Agent:} We build a budget-decoupled navigation environment with unrestricted parent re-entry and JevNavAgent, pairing three factorized single-forward heads (evidence-sufficiency stopping, spatial-geometry candidate scoring, and gated contextual-bilinear answering) with magnification context and hierarchical overview modality dropout. Together, the gated contextual-bilinear head, magnification context, and overview modality dropout jointly lower Step-0/Step-$T$ rigidity from $98.58\%$ to $58.69\%$ so that answers depend on newly zoomed evidence.
    \item \textbf{Counterfactual Trajectory Alignment (Stage 1):} We formalize the Existential-vs.-Universal Stopping Asymmetry and synthesize depth- and kinematics-aligned counterfactual multi-branch and margin-pan trajectories with stop-decoupled answer supervision, substantially mitigating step-count and panning shortcuts and enabling zero-shot test-time budget scaling over $B \in \{4, 6, 8, 10, 12\}$.
    \item \textbf{Cost-Effective Online Agentic RL (Stage 2):} Exploiting single-forward rollouts, we introduce trajectory-level GRPO~\cite{shao2024deepseekmath} with three levels of credit: branch-level lesion discovery, in-polygon immediate stopping against hard-negative follicles, and step-wise answer-confidence gain ($\Delta P_t$) under a soft KL lock to the Stage-1 policy.
    \item \textbf{Two-Benchmark Validation:} Against $16$ baselines and controls on Clean Visual WSI-VQA and $10$ on CAMELYON16, JevNavAgent reaches $51.57\%$ multi-scale ($50.43\%$ Step-$T$) accuracy on WSI-VQA; on the CAMELYON16 Official Test ($B=12$), Traj-GRPO Online achieves the highest Strict Loc+Spec ($67.44\%$), while Traj-GRPO Calib.\ achieves the highest localization ($22.45\%$) and slide accuracy ($69.77\%$) among all compared methods.
\end{enumerate}

\section{Related Work}
\label{sec:related}

\subsection{Gigapixel WSI Representation and Multiple Instance Learning}
Because gigapixel WSIs cannot fit into GPU memory at native resolution, standard computational pathology pipelines separate patch encoding from slide-level aggregation under weak supervision~\cite{campanella2019clinical}. Attention-based MIL (ABMIL)~\cite{ilse2018attention} introduces attention-based pooling over instance embeddings, CLAM~\cite{lu2021data} adds instance-level clustering constraints, and TransMIL~\cite{shao2021transmil} and HIPT~\cite{chen2022scaling} model inter-patch spatial correlations and hierarchical tissue representations. Recent pathology foundation models substantially strengthen these pipelines: CTransPath~\cite{wang2022transformer} and UNI~\cite{chen2024uni} learn transferable visual encoders, CONCH~\cite{lu2024visual} aligns histopathology images with text, and Prov-GigaPath~\cite{xu2024whole} combines tile-level and whole-slide pretraining. Nevertheless, these paradigms operate over offline pre-extracted, fixed slide evidence rather than performing online, query-conditioned visual acquisition of new high-magnification fields in response to intermediate findings.

\subsection{Pathology Vision-Language Models and Whole-Slide VQA}
Recent pathology and medical VLMs support diverse diagnostic interfaces on localized regions of interest (ROIs). PLIP~\cite{huang2023visual} learns contrastively aligned image--text representations for retrieval and zero-shot classification, whereas Quilt-LLaVA~\cite{seyfioglu2024quilt}, PathChat~\cite{lu2024multimodal}, and the general medical MedGemma family~\cite{sellergren2025medgemma} enable instruction-following visual dialogue. Patho-R1~\cite{zhang2025pathor1} further applies reinforcement learning to enhance generated diagnostic reasoning over pathology images. To move beyond pre-cropped ROIs, WSI-VQA~\cite{chen2024wsivqa} generates answers over slide-level patch features via a Wsi2Text Transformer, and SlideChat~\cite{chen2024slidechat} connects whole-slide representations to a large language assistant. However, answering from precomputed tile embeddings still differs from actively deciding where to zoom. When general or pathology VLMs are evaluated directly on whole slides without oracle ROI crops or exhaustive patch pre-extraction, relying on a low-resolution whole-slide overview leaves fine cellular detail illegible and often induces affirmative bias on unresolved focal findings (Section~\ref{subsec:exp_wsivqa}).

\subsection{Active Visual Navigation, Single-Forward Decisions, and Agentic RL}
Selective visual inspection, which sequentially chooses where to look in a high-resolution scene, has roots in recurrent hard attention~\cite{mnih2014recurrent,ba2014multiple} and reinforcement learning over image grids and scale spaces~\cite{uzkent2020efficient,papadopoulos2021hard}. In computational pathology, two representative paradigms are: (i) retrieval and scanning pipelines, including contrastive cascades such as PLIP~\cite{huang2023visual} with a terminal VLM and PathNavigate~\cite{yang2026pathnavigate}, which combines slide memory with surprise-guided scanning and query-conditioned reranking; and (ii) multi-turn autoregressive VLM agents such as PathAgent~\cite{chen2025pathagent}, which coordinate multi-step exploration, generated reflection, and termination across dialogue turns. While autoregressive agents can revise their search dynamically, decoding action strings and natural-language reasoning at every step incurs substantial latency; in our unified-environment benchmark, we adapt both PathNavigate and PathAgent to a shared candidate pool and fixed step budget for controlled comparison.

TypeSafe AI's Jev paradigm~\cite{almeida2026jev,typesafe2026documentation} offers a distinct ``System One'' decision interface inspired by fast intuitive judgment~\cite{kahneman2011thinking}. Rather than generating natural-language explanations token by token, a Jev model reads context and directly outputs probability distributions over predefined categorical, ordinal, or binary decision spaces, while deterministic code orchestrates the surrounding workflow. We adapt this principle to multi-step gigapixel navigation on a Qwen3-VL-2B backbone~\cite{bai2025qwen3vl}: at each step, a single multimodal forward pass produces factorized stopping, candidate-view, and answer distributions with zero generated tokens, while deterministic code manages quadtree candidate generation, unrestricted parent re-entry, and root switching.

Training such a multi-step inspection policy requires overcoming both imitation shortcuts and trajectory-level credit assignment. Offline behavioral cloning is vulnerable to state-distribution shift (addressed by DAgger~\cite{ross2011reduction}), causal confusion~\cite{de2019causal}, and shortcut learning~\cite{geirhos2020shortcut}. In WSI screening, these failures manifest as spurious correlations between slide labels and trajectory depth, panning transitions, or global thumbnails due to the asymmetry between existential positive detection and universal negative exclusion. Meanwhile, recent pathology RL work such as Patho-R1~\cite{zhang2025pathor1} applies Group Relative Policy Optimization (GRPO)~\cite{shao2024deepseekmath} to single-turn generative reasoning, whereas extending online RL to multi-step visual acquisition and stopping is computationally costly under autoregressive token decoding. By eliminating decoding overhead, JevNavAgent makes online multi-trajectory rollouts practical, pairing Stage-1 counterfactual trajectory alignment with Stage-2 trajectory-level GRPO for visual acquisition, branch discovery, and calibrated stopping.

\section{Methodology}
\label{sec:method}

\begin{figure*}[t]
\centering
\includegraphics[width=0.98\textwidth]{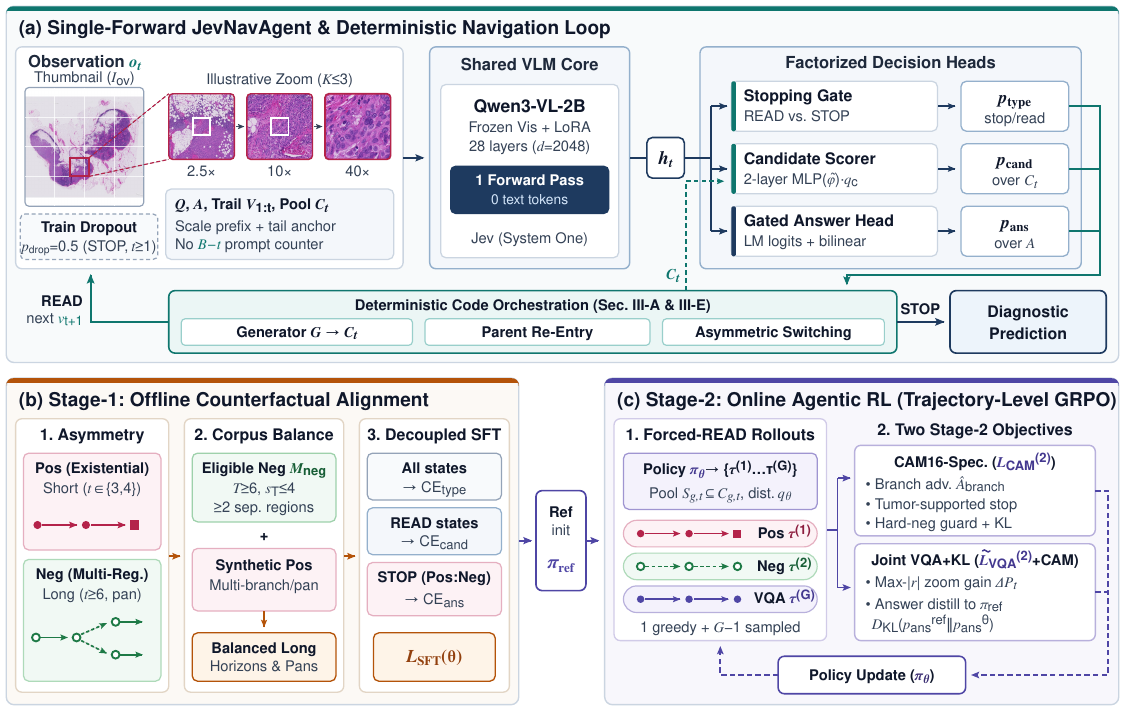}
\caption{Overview of the JevNavAgent architecture and two-stage trajectory alignment framework. \textbf{(a)} Deterministic environment code supplies a multi-scale observation $o_t$ (whole-slide thumbnail $I_{\text{ov}}$, up to $K=3$ recent crops shown with illustrative nested $2.5\times\text{--}40\times$ fields from a CAMELYON16 Official Test episode, query $Q$, options $\mathcal{A}$, trail $V_{1:t}$, and candidate pool $C_t$, without $B-t$ prompt counters) to Qwen3-VL-2B-Instruct. Under the Jev (System One) interface, one forward pass ($0$ generated tokens) yields $p_{\text{type}}$, $p_{\text{cand}}$ (scoring $C_t$), and $p_{\text{ans}}$ (reusing shared LM logits with bilinear keys); deterministic orchestration executes $\text{READ}$ (next view $v_{t+1}$) or $\text{STOP}$ (final diagnostic prediction). \textbf{(b)} Stage-1 SFT offline counterfactual alignment balances long-horizon and panning stops via eligible multi-region negatives ($\mathcal{M}_{\text{neg}}$), synthetic positives, and stop-decoupled supervision ($\mathcal{L}_{\text{SFT}}$). \textbf{(c)} Stage-2 Traj-GRPO rolls out forced-$\text{READ}$ groups over restricted pools $\mathcal{S}_{g,t}$ for two specializations: + GRPO (CAM16) ($\mathcal{L}_{\text{CAM}}^{(2)}$) and + GRPO (VQA+KL) ($\tilde{\mathcal{L}}_{\text{VQA}}^{(2)}+\text{CAM}$).}
\label{fig:method_overview}
\end{figure*}

We formulate gigapixel WSI interpretation as an active sequential decision process over a multi-resolution image pyramid. As illustrated in Fig.~\ref{fig:method_overview}, our methodology comprises five components: (1) a partially observable Markov decision process (POMDP) with deterministic multi-scale candidate generation and unrestricted parent re-entry (Section~\ref{subsec:pomdp}); (2) JevNavAgent, a single-forward decision architecture inspired by the Jev (System One) paradigm that maps multimodal observations directly to stopping, spatial-candidate, and answer distributions without autoregressive text generation (Section~\ref{subsec:architecture}); (3) Stage-1 offline counterfactual trajectory alignment to mitigate horizon and kinematic shortcuts (Section~\ref{subsec:debiasing}); (4) Stage-2 online agentic reinforcement learning via trajectory-level GRPO (Section~\ref{subsec:traj_grpo}); and (5) an asymmetric multi-region inference and test-time budget scaling protocol (Section~\ref{subsec:inference}).

\subsection{POMDP Formulation and Deterministic Multi-Scale Environment}
\label{subsec:pomdp}

\subsubsection{Multi-Scale Pyramid and Bounded Observation State}
Let $\mathcal{W}$ denote a gigapixel whole-slide image of level-0 pixel dimensions $(W, H)$ queried over discrete pyramid downsample factors $\mathcal{S} = \{64, 16, 4, 1\}$ (corresponding approximately to $0.625\times$, $2.5\times$, $10\times$, and $40\times$ magnifications at $0.25\,\mu\text{m/pixel}$). Each acquired field of view (FOV) at step $t \ge 1$ is parameterized by
\begin{equation}
    \label{eq:fov_tuple}
    v_t = (x_t, y_t, s_t, b_t), \quad (x_t, y_t) \in [0, 1]^2, \; s_t \in \mathcal{S},
\end{equation}
where $(x_t, y_t)$ is the normalized center on the level-0 canvas and $b_t = [x_{t,0}, y_{t,0}, x_{t,1}, y_{t,1}]$ is the level-0 pixel bounding box of a $512 \times 512$ crop extracted at scale $s_t$ (white-padded if $b_t$ extends past slide borders).

At step $t \ge 0$, given clinical query $Q$ and option list $\mathcal{A} = (a_1, \dots, a_M)$ (typically $M=4$ for WSI-VQA, with binary $M=2$ cases also supported, and $M=2$ for metastasis screening), the policy observes the public state summarized by
\begin{equation}
    \label{eq:obs_state}
    o_t = \Big( I_{\text{ov}}, \; \{I_k\}_{k=\max(1, t-K+1)}^{t}, \; V_{1:t}, \; C_t, \; Q, \; \mathcal{A} \Big),
\end{equation}
alongside slide geometry metadata (pixel dimensions and microns-per-pixel). Here $I_{\text{ov}}$ is the whole-slide overview thumbnail (extracted up to $1024 \times 1024$ pixels and encoded under a per-image visual pixel cap of $262{,}144$ pixels), $\{I_k\}$ is empty at $t=0$ and retains at most the $K=3$ most recent $512 \times 512$ crops for $t \ge 1$, $V_{1:t} = (v_1, \dots, v_t)$ is the symbolic view trail, and $C_t = (c_1, \dots, c_{N_t})$ is the navigable candidate list ($C_t = \emptyset$ once $t = B$). Capping the crop buffer at $K=3$ bounds per-step visual encoding cost, while symbolic history length grows linearly with $t$.

\subsubsection{Candidate Generation and Unrestricted Parent Re-Entry}
Following the Jev separation between neural judgment and deterministic workflow code~\cite{almeida2026jev,typesafe2026documentation}, a deterministic generator $\mathcal{G}(V_{1:t}, I_{\text{ov}})$ constructs $C_t$ ($N_t \le 32$ when $t < B$):
\begin{itemize}
    \item Sub-Quadrant Refinement ($t \ge 1$, $s' = \max(1, s_t/4)$): A $3 \times 3$ sub-grid around $v_t$, spaced by $512\,s'$ level-0 pixels when $s_t > 1$ and by quarter-field $128$-pixel offsets when $s_t = 1$.
    \item Cardinal Local Panning ($t \ge 1$, $s' = s_t$): Four cardinal neighbors shifted by half-field strides ($256\,s_t$ level-0 pixels) at scale $s_t$ to track margins or partially clipped lesions.
    \item Cross-Region Root Proposals ($t \ge 0$, $s' = 64$): Up to $N_{\text{root}} = 16$ stratified root centers across a $4 \times 4$ thumbnail grid, snapped to the nearest foreground pixel in each cell using an RGB tissue mask ($\max_c - \min_c > 18$, $\min_c < 235$, $\max_c > 40$), with a uniform $4 \times 4$ cell-center fallback if no foreground pixel is detected.
\end{itemize}

A history-wide visited-view mask would lock coarse parent nodes ($s \in \{64, 16\}$) after one visit and block access to their remaining sub-quadrants. Under Unrestricted Parent Re-Entry, $\mathcal{G}$ deduplicates proposals within $C_t$ by quantized box key $(x_0, y_0, s)$ and masks only the current standing view $v_t$, leaving previously visited coarse parents accessible for sub-branch exploration.

\subsubsection{Budget-Decoupled Policy Conditioning}
Explicit remaining-step counters $(B - t)$ in the policy prompt can encourage policies to rely on counter exhaustion as a termination cue. The environment tracks $B - t$ externally, checking whether remaining budget permits another read ($m_t^{\text{read}} = \mathbb{I}[N_t > 0 \wedge B - t > 0]$) and setting $C_B = \emptyset$ at the safety horizon. We omit remaining-step fields when serializing the text prompt and delegate horizon enforcement to an external execution wrapper.

\subsection{JevNavAgent: Single-Forward Decision Architecture}
\label{subsec:architecture}

Autoregressive VLM agents emit coordinates and reasoning across dozens of output tokens per step, accumulating multi-turn decoding latency. Drawing on the Jev (System One) paradigm~\cite{almeida2026jev,typesafe2026documentation}, we instantiate JevNavAgent on Qwen3-VL-2B-Instruct~\cite{bai2025qwen3vl} ($28$ language layers, hidden size $d = 2048$) with frozen visual weights and language-layer LoRA adapters ($r=16, \alpha=32$, dropout $0.05$). At each step $t$, one multimodal forward pass ($0$ generated tokens) produces distributions over (i) $\{\text{READ}, \text{STOP}\}$, (ii) $C_t$ (when $N_t > 0$), and (iii) $\mathcal{A}$.

\subsubsection{Scale-Aware Prompt Recency and Tail Re-Anchoring}
To preserve magnification context and question--option alignment near the readout position, we format the user message with two recency cues: (1) Explicit Magnification Prefixing, prepending a header with $s_k$ and rounded center $(x_k, y_k)$ before each crop $I_k$; and (2) Tail Re-Anchoring, appending the visited downsample list $(s_1, \dots, s_t)$, question $Q$, and letter-indexed options $\{\text{``A''}: a_1, \dots\}$ at the end of user content. Under the multimodal chat template, let $H_{\text{last}} \in \mathbb{R}^{d}$ denote the final-layer hidden state at the terminal non-padding prompt token and $h_t = \text{LayerNorm}(H_{\text{last}}) \in \mathbb{R}^{d}$.

\subsubsection{Factorized Decision Heads}
Three lightweight heads ($d_{\text{proj}} = 256$) read out from $h_t$:

\paragraph{Evidence-Sufficiency Stopping Gate}
A linear classifier outputs exploration ($\text{READ}=0$) versus termination ($\text{STOP}=1$) probabilities:
\begin{equation}
    \label{eq:action_head}
    p_{\text{type}}\!\left(a_t^{\text{type}} \mid o_t\right) = \text{Softmax}\!\left(W_{\text{type}} h_t + b_{\text{type}}\right),
\end{equation}
with $W_{\text{type}} \in \mathbb{R}^{2 \times d}$ and $b_{\text{type}} \in \mathbb{R}^2$.

\paragraph{Spatial-Geometry Candidate Scorer}
For $N_t > 0$, each candidate $c_i \in C_t$ carries a 12-dimensional vector $\phi(c_i) \in \mathbb{R}^{12}$: normalized center $(x_i, y_i)$, $\log_4(s_i)/3$, normalized box $(x_{i,0}/W, y_{i,0}/H, x_{i,1}/W, y_{i,1}/H)$, thumbnail tissue fraction $\tau(c_i) \in [0, 1]$, FOV-normalized center offsets $\Delta x_i = (W x_i - \text{cx}(b_t))/\text{width}(b_t)$ and $\Delta y_i = (H y_i - \text{cy}(b_t))/\text{height}(b_t)$ ($0$ at $t=0$), stratified-root operator flag $r_i = \mathbb{I}[c_i \text{ from root grid}]$ (distinct from $s_i = 64$, as DS-64 pans have $r_i = 0$), and relative scale step $\log_4(s_t / s_i)/3$ ($0$ at $t=0$). While serialized prompt metadata retain raw $\phi(c_i)$, the neural scorer applies $\tanh$ to $(\Delta x_i, \Delta y_i)$ to form $\tilde{\phi}(c_i)$, then scores candidates via a two-layer GELU MLP $\psi_{k,c}: \mathbb{R}^{12} \to \mathbb{R}^{d_{\text{proj}}}$ and linear query $q_c(h_t) = W_{q,c} h_t + b_{q,c}$:
\begin{align}
    s_{\text{cand}}(c_i \mid o_t) &= \frac{\psi_{k,c}\!\big(\tilde{\phi}(c_i)\big)^\top q_c(h_t)}{\sqrt{d_{\text{proj}}}}, \label{eq:cand_score} \\
    p_{\text{cand}}(c_i \mid o_t) &= \frac{\exp\!\big(s_{\text{cand}}(c_i \mid o_t)\big)}{\sum_{j=1}^{N_t} \exp\!\big(s_{\text{cand}}(c_j \mid o_t)\big)}. \label{eq:cand_head}
\end{align}

\paragraph{Gated Contextual-Bilinear Answer Head}
Let $e(a_j) \in \mathbb{R}^d$ be the masked mean of Layer-0 input token embeddings for option string $a_j$, and let $\nu_j$ be the vocabulary token ID of option letter $j \in \{\text{``A''}, \text{``B''}, \dots\}$. While a bilinear score between contextual query $h_t$ and static option keys $e(a_j)$ uses all 28 layers on the query side, its option representation remains context-free; we therefore blend it with pretrained language-head letter logits $\ell_{\text{LM}}(\nu_j) = \text{LMHead}(H_{\text{last}})[\nu_j]$ via an instance-dependent gate $g_t \in (0, 1)$:
\begin{align}
    z_{\text{bil}}(a_j \mid o_t) &= \frac{\big(W_{k,a}\,e(a_j) + b_{k,a}\big)^\top \big(W_{q,a}\,h_t + b_{q,a}\big)}{\sqrt{d_{\text{proj}}}}, \label{eq:z_bil} \\
    z_{\text{ans}}(a_j \mid o_t) &= (1 - g_t)\,z_{\text{bil}}(a_j \mid o_t) + g_t\,\tilde{\ell}_{\text{LM}}(\nu_j), \label{eq:z_ans}
\end{align}
where $\tilde{\ell}_{\text{LM}}(\nu_j) = \ell_{\text{LM}}(\nu_j) - \frac{1}{M}\sum_{m=1}^{M} \ell_{\text{LM}}(\nu_m)$ (mean-centering is shift-invariant under the shared gate $g_t$ and $\text{Softmax}$), $g_t = \sigma(w_g^\top h_t + b_g)$ ($w_g \in \mathbb{R}^d$ and $b_g \in \mathbb{R}$ zero-initialized so $g_t = 0.5$ initially), and $p_{\text{ans}}(a_j \mid o_t) = \text{Softmax}_j(z_{\text{ans}}(a_j \mid o_t))$.

\subsubsection{Stage-1 Hierarchical Overview Modality Dropout}
Because $I_{\text{ov}}$ appears at index 0 in every state, our M0 baseline ($z_{\text{bil}}$ without prompt recency or dropout) exhibits $98.58\%$ Step-0/Step-$T$ prediction agreement, consistent with heavy reliance on overview cues. To encourage reliance on acquired crops, during Stage-1 SFT we apply Hierarchical Overview Modality Dropout on terminal $\text{STOP}$ states with at least one crop ($t \ge 1$), replacing $I_{\text{ov}}$ with a $384 \times 384$ RGB $(210, 210, 210)$ neutral gray image $I_{\text{gray}}$ with probability $p_{\text{drop}} = 0.5$:
\begin{equation}
    \label{eq:mod_dropout}
    \begin{aligned}
        \tilde{I}_{\text{ov}} &=
        \begin{cases}
            I_{\text{gray}}, & \xi = 1, \\
            I_{\text{ov}}, & \xi = 0,
        \end{cases} \\
        \xi &\sim \text{Bernoulli}\!\left(p_{\text{drop}} \cdot \mathbb{I}\!\left[a_t^{\text{type}*} = \text{STOP} \wedge t \ge 1\right]\right).
    \end{aligned}
\end{equation}
Intermediate $\text{READ}$ states and Stage-2 rollouts always retain $I_{\text{ov}}$, preserving global spatial context for candidate navigation while curbing thumbnail over-reliance during terminal answer readout.

\subsection{Stage-1 Offline Counterfactual Trajectory Alignment}
\label{subsec:debiasing}

\subsubsection{Multi-Scale Trajectory Construction Across Tasks}
Let $y^* \in \{1, \dots, M\}$ denote the ground-truth option index and $a^* = a_{y^*}$ its option text. We construct offline trajectories teaching candidate selection, recovery, and stopping across both benchmarks:
\begin{itemize}
    \item Scale-Aware Tree Search and Recovery for WSI-VQA: Without region annotations, we rank quadtree candidates against $(Q, a^*)$ using PLIP~\cite{huang2023visual} under scale-dependent actual-pixel foreground gates ($0.03 / 0.12 / 0.25 / 0.25$ for $\text{DS} \in \{64, 16, 4, 1\}$, with a coarse-root fallback if needed). Macro queries generate up to two routes targeting $\text{DS}=4$; focal and grading queries generate up to three routes targeting $\text{DS}=1$ (with a $\text{DS}=4$ fallback when no qualified $\text{DS}=1$ route exists), adding a DAgger-inspired~\cite{ross2011reduction} distractor-recovery route from a low-ranked tissue root whose deliberately perturbed initial step is excluded from imitation targets while retaining recovery states.
    \item Annotation-Guarded VLM and Synthetic Routes for Screening: On CAMELYON16 training slides, we combine candidate preferences from a frontier VLM (GPT-6.1-Terra) collected under annotation-guarded stopping and oracle-closed positive chains with synthetic multi-branch and near-miss recovery routes. Ground-truth polygons are used strictly during offline construction, Stage-2 reward computation, and evaluation.
\end{itemize}

\subsubsection{Stopping Asymmetry and Counterfactual Corpus Balancing}
Clinical screening exhibits a structural asymmetry: confirming tumor is existential (one verified high-magnification field suffices, permitting short $t \in \{3, 4\}$ trajectories), whereas ruling out tumor across suspicious regions requires multi-region inspection ($t \ge 6$ and frequent lateral pans). In unaligned corpora, these systematic differences in stopping depth, horizon length $t$, and terminal three-step scale/panning patterns $(s_{t-2}, s_{t-1}, s_t)$ leak slide-level labels, allowing a student policy to exploit non-visual trajectory kinematics as shortcuts instead of cellular evidence. We mitigate this in Stage-1 corpus $\mathcal{D}_{\text{train}}$ via two steps:

\paragraph{Admissible Multi-Region Negative Stops}
Every CAMELYON16 negative terminal record $(o_T, \text{STOP}, Y=0)$ is restricted to
\begin{equation}
    \label{eq:manifold_neg}
    \mathcal{M}_{\text{neg}} = \big\{ V_{1:T} \;\big|\; T \ge 6, \, s_T \in \{4, 1\}, \, |\mathcal{R}_{8000}(V_{1:T})| \ge 2 \big\},
\end{equation}
where $|\mathcal{R}_{8000}(V_{1:T})|$ is the greedy count of visited $\text{DS} \le 4$ centers separated by $\max(|\Delta X|, |\Delta Y|) \ge 8{,}000$ level-0 pixels. Negative source routes are truncated to up to three eligible prefixes (first/last odd and last even horizons in $\{6, \dots, 12\}$) satisfying $\mathcal{M}_{\text{neg}}$, or discarded otherwise.

\paragraph{Long-Horizon and Pan/Refine Equalization}
On positive CAMELYON16 slides, alongside $19$ retained early stops ($t \in \{3, 4\}$), we include synthetic routes with benign prefixes of varying length ($0\text{--}8$ steps) and near-miss margin recovery (via panning or refinement), then equalize empirical positive and negative terminal record counts on the long-horizon CAMELYON16 terminal subset $\mathcal{D}_{\text{CAM}}^{\text{long}}$ across every horizon $t \in \{6, \dots, 12\}$ and terminal scale $s_T \in \{4, 1\}$:
\begin{equation}
    \label{eq:inv_horizon}
    \begin{aligned}
        P_{\mathcal{D}_{\text{CAM}}^{\text{long}}}\!\big(Y = 1 \;\big|\; a_t^{\text{type}*} = \text{STOP}, \, t, \, s_T\big) &= 0.5, \\
        \forall\, t \in \{6, \dots, 12\}, \; s_T &\in \{4, 1\}.
    \end{aligned}
\end{equation}
This reduces imbalance on dominant terminal 3-step patterns, yielding positive-to-negative stop counts of $113{:}125$ on $(64,16,4)$, $26{:}23$ on $(16,4,1)$, and $12{:}13$ on lateral pans $(16,4,4)$. We also upsample Center-1 negative $\text{READ}$ states and focal affirmative VQA stops to reduce center and answer skew.

\subsubsection{Stop-Decoupled Supervision Objective}
In $\mathcal{D}_{\text{train}}$, intermediate $\text{READ}$ records account for $87.5\%$ ($2{,}450/2{,}799$) of CAM records and carry a negative field label prior to reaching tumor; supervising $p_{\text{ans}}$ on intermediate $\text{READ}$ steps suppresses positive sensitivity (evaluated as the ``READ-Neg'' ablation in Section~\ref{sec:experiments}). Our Stage-1 SFT policy instead uses stop-decoupled supervision, training $p_{\text{cand}}$ exclusively on $\text{READ}$ states and $p_{\text{ans}}$ exclusively on class-balanced $\text{STOP}$ states:
\begin{equation}
    \label{eq:total_loss}
    \begin{aligned}
        \mathcal{L}_{\text{SFT}}(\theta) &= \mathcal{L}_{\text{CE}}\!\left(p_{\text{type}}(o_t), \, a_t^{\text{type}*}\right) \\
        &\quad + \mathbb{I}\!\left[a_t^{\text{type}*} = \text{READ}\right] \mathcal{L}_{\text{CE}}\!\left(p_{\text{cand}}(o_t), \, c_t^*\right) \\
        &\quad + \mathbb{I}\!\left[a_t^{\text{type}*} = \text{STOP}\right] \mathcal{L}_{\text{CE}}\!\left(p_{\text{ans}}(o_t), \, y^*\right),
    \end{aligned}
\end{equation}
where $c_t^*$ is the target candidate index.

\subsection{Stage-2 Online Agentic RL via Trajectory-Level GRPO}
\label{subsec:traj_grpo}

State-wise SFT (Eq.~\eqref{eq:total_loss}) lacks trajectory-level credit assignment: on held-out slides, it can visit a micrometastasis at $s_t \le 4$ (``Visited Hit'') yet continue reading back into benign tissue. Because JevNavAgent requires $0$ generated tokens per step, rolling out online trajectory groups $\{\tau^{(g)}\}_{g=1}^G$ avoids autoregressive decoding overhead. Rollouts execute forced $\text{READ}$ steps up to a fixed rollout horizon over a structured root/refinement candidate subset $\mathcal{S}_{g,t} \subseteq C_{g,t}$ (one greedy trajectory per group and $G-1$ trajectories sampled at temperature $0.85$), and updates evaluate the restricted candidate distribution $q_\theta(c \mid o_{g,t}, \mathcal{S}_{g,t}) = \exp s_{\text{cand}}(c \mid o_{g,t}) / \sum_{c' \in \mathcal{S}_{g,t}} \exp s_{\text{cand}}(c' \mid o_{g,t})$ directly. Initialized from the Stage-1 SFT reference policy $\pi_{\text{ref}}$, we study two Stage-2 Traj-GRPO specializations (analyzed in Table~\ref{tab:grpo_ablation}): + GRPO (CAM16) ($3$ positive and $2$ negative $B{=}12$ rollouts per update) for gigapixel screening, and + GRPO (VQA+KL) ($2$ WSI $B{=}3$ rollouts plus $2$ positive and $1$ negative CAM $B{=}6$ rollouts per update) for multi-choice VQA.

\subsubsection{CAM16 Trajectory-Level Branch and Stopping Objective}
Each positive rollout $g$ is partitioned into 3-step coarse-to-fine branches $b$ ending at leaf view $v_{\text{leaf}}$. Let $f_{\text{win}}(v)$ denote the tumor fraction inside the central $512 \times 512$ level-0 scoring window of view $v$, and let $\text{Hit}(v)$ denote a qualified high-resolution hit (center inside tumor polygon, $f_{\text{win}}(v) \ge 0.25$, in-bounds window, and $s \le 4$). Across all positive-slide branches $\mathcal{B}$, each branch receives reward and critic-free group-normalized advantage~\cite{shao2024deepseekmath}:
\begin{align}
    r_{\text{branch}}(b) &= 1.5\,\mathbb{I}\!\left[\text{Hit}(v_{\text{leaf}})\right] + 0.8 \max_{v_k \in b} f_{\text{win}}(v_k), \label{eq:branch_reward} \\
    \hat{A}_{\text{branch}}(b) &= \frac{r_{\text{branch}}(b) - \mu_{\mathcal{B}}}{\sigma_{\mathcal{B}} + \varepsilon}\,\mathbb{I}\!\left[\sigma_{\mathcal{B}} > \varepsilon\right], \label{eq:branch_adv}
\end{align}
with sample standard deviation $\sigma_{\mathcal{B}}$ and $\varepsilon = 10^{-5}$. On active rollout-step pairs $u = (g, t)$, the + GRPO (CAM16) objective combines three credit mechanisms:
\begin{itemize}
    \item Branch-Level Discovery Credit ($\ell_{\text{br},u}$): On readable steps with $|\hat{A}_{\text{branch}}(b(u))| > \delta_{\text{adv}}$ ($\delta_{\text{adv}} = 0.15$), $\ell_{\text{br},u} = -\alpha_b\,\hat{A}_{\text{branch}}(b(u))\log q_\theta(c_u \mid o_u, \mathcal{S}_u)$ ($\alpha_b = 0.35$; $0$ otherwise), rewarding branches that reach tumor even if subsequent steps fail to stop.
    \item Tumor-Supported Stopping Alignment ($\ell_{\text{align},u}$): At high magnification ($s_u \le 4, t > 0$) on positive rollouts, tumor-supported views ($\text{Hit}(v_u) \vee f_{\text{win}}(v_u) \ge 0.30$) incur $-1.6(1 + 0.5 f_{\text{win}}(v_u))(\log p_{\text{type}}(\text{STOP}\mid o_u) + \log p_{\text{ans}}(\text{Pos}\mid o_u))$, whereas zero-overlap benign leaves ($f_{\text{win}}(v_u) = 0$) incur $-0.65\,m_u^{\text{read}}\log p_{\text{type}}(\text{READ}\mid o_u) - 0.55\,\mathbb{I}[p_{\text{ans}}(\text{Pos}\mid o_u) > 0.5]\log p_{\text{ans}}(\text{Neg}\mid o_u)$ (and $0$ otherwise).
    \item Hard-Negative Follicle Guard ($\ell_{\text{neg},u}$) and Replay: Across negative rollouts, $\mathcal{H}_-$ selects the top two $s_u \le 4$ states per rollout with highest false-positive risk $p_{\text{stop}}(o_u)p_{\text{pos}}(o_u)$, applying $\ell_{\text{neg},u} = -0.80(m_u^{\text{read}}\log p_{\text{type}}(\text{READ}\mid o_u) + \log p_{\text{ans}}(\text{Neg}\mid o_u))$, alongside one offline positive-hit anchor $\ell_{\text{hit}}$ (weight $1.50$ plus KL) and one offline WSI-VQA replay state $\ell_{\text{WSI}} = \mathcal{L}_{\text{SFT}}$.
\end{itemize}
Averaged over $N_{\text{terms}}$ active state contributions (where $\mathcal{J}_+$ denotes positive states with $(s_u \le 4 \wedge t > 0) \vee |\hat{A}_{\text{branch}}(b(u))| > \delta_{\text{adv}}$), the + GRPO (CAM16) loss is
\begin{equation}
    \label{eq:cam_grpo}
    \begin{aligned}
        \mathcal{L}_{\text{CAM}}^{(2)}(\theta) &= \frac{1}{N_{\text{terms}}}\Bigg[ \sum_{u \in \mathcal{J}_+}\!\big(\ell_{\text{br},u} + \ell_{\text{align},u} + \beta_c d_u\big) \\
        &\quad + \sum_{u \in \mathcal{H}_-}\!\big(\ell_{\text{neg},u} + \beta_c d_{\text{ta}}(o_u)\big) + \ell_{\text{hit}} + 0.50\,\ell_{\text{WSI}} \Bigg],
    \end{aligned}
\end{equation}
where $\beta_c = 0.05$, $d_{\text{ta}}(o) = D_{\text{KL}}(p_{\text{type}}^\theta \,\|\, p_{\text{type}}^{\text{ref}})_o + D_{\text{KL}}(p_{\text{ans}}^\theta \,\|\, p_{\text{ans}}^{\text{ref}})_o$, and $d_u = d_{\text{ta}}(o_u) + \mathbb{I}[\ell_{\text{br},u} \neq 0]D_{\text{KL}}(q_\theta \,\|\, q_{\text{ref}})_u$.

\subsubsection{Joint VQA+KL Confidence-Gain and Distillation Objective}
On WSI-VQA rollout $\tau^{(g)}$, each zoom transition $o_{g,t} \to o_{g,t+1}$ receives reward $r_{g,t} = 2\Delta P_{g,t} + r_{\text{fg}}(I_{g,t+1})$, where $\Delta P_{g,t} = p_{\text{ans}}(y^* \mid o_{g,t+1}) - p_{\text{ans}}(y^* \mid o_{g,t})$ and $r_{\text{fg}}(I_{g,t+1}) = +0.20$ if crop foreground $\ge 0.25$ (else $-0.40$). Let $\text{IG}_g = p_{\text{ans}}(y^* \mid o_{g,T}) - p_{\text{ans}}(y^* \mid o_{g,0})$ be the detached rollout gain, $\hat{A}_g$ the group-normalized advantage of episode return $R_g = \mathbb{I}[\hat{y}_{g,T}{=}y^*] + 0.8 p_{\text{ans}}(y^*\mid o_{g,T}) + 1.2\,\text{IG}_g + r_{\text{scale}} - 0.3\,\mathbb{I}[\text{fg}_{g,T} < 0.15]$ ($r_{\text{scale}} = 0.15$ on $B{=}3$ rollouts), and $t_g^* = \arg\max_t |r_{g,t}|$ the single highest-magnitude zoom transition per rollout. With $w_V = 1.2$ and $\beta_v = 0.06$, the VQA contribution is
\begin{equation}
    \label{eq:vqa_grpo}
    \begin{aligned}
        \tilde{\mathcal{L}}_{\text{VQA}}^{(2)}(\theta) &= \sum_{g=1}^{G} \Big[ -w_V\big(0.45 r_{g,t_g^*} + 0.25\hat{A}_g\big)\log q_{\theta,g}(c_{g,t_g^*}) \\
        &\quad + \beta_v D_{\text{KL}}\!\big(q_\theta \,\|\, q_{\text{ref}}\big)_{g,t_g^*} \\
        &\quad + 1.5 w_V D_{\text{KL}}\!\big(p_{\text{ans}}^{\text{ref}} \,\|\, p_{\text{ans}}^\theta\big)_{g,T} \\
        &\quad - 0.25 w_V \mathbb{I}[\text{IG}_g > 0]\log p_{\text{ans}}^\theta(y^* \mid o_{g,T}) \Big] \\
        &\quad + 1.5 w_V D_{\text{KL}}\!\big(p_{\text{ans}}^{\text{ref}} \,\|\, p_{\text{ans}}^\theta\big)_{\text{offline}},
    \end{aligned}
\end{equation}
where $q_{\theta,g}(c_{g,t}) = q_\theta(c_{g,t} \mid o_{g,t}, \mathcal{S}_{g,t})$ and forward distillation $D_{\text{KL}}(p_{\text{ans}}^{\text{ref}} \,\|\, p_{\text{ans}}^\theta)$ on terminal and offline WSI states discourages multi-class answer-distribution drift. In + GRPO (VQA+KL), $\tilde{\mathcal{L}}_{\text{VQA}}^{(2)}$ is optimized jointly with a companion screening term $\tilde{\mathcal{L}}_{\text{CAM,joint}}^{(2)}$ ($\alpha_b = 0.45, \delta_{\text{adv}} = 0.20$ on initial branch steps, tumor-supported stopping alignment at $f_{\text{win}} \ge 0.25$, and adaptive hard-negative guarding), normalized by total active updates.

\subsection{Asymmetric Multi-Region Inference and Test-Time Budget Scaling}
\label{subsec:inference}

\subsubsection{Asymmetric Screening Orchestration and Trace Readout}
\label{subsubsec:screening_orchestration}
Let $p_{\text{stop}}(o_t) = p_{\text{type}}(\text{STOP} \mid o_t)$, $p_{\text{pos}}(o_t) = p_{\text{ans}}(\text{Pos} \mid o_t)$, and $H_t$ be the greedy count of visited $s \le 4$ centers separated by normalized Euclidean distance $\ge 0.06$:
\begin{enumerate}
    \item Online Multi-Region Orchestration: At $s_t \le 4$, the positive evidence criterion is $\mathcal{E}_+(o_t) = \mathbb{I}[(p_{\text{stop}} \ge 0.50 \wedge p_{\text{pos}} \ge 0.885) \vee (H_t > 1 \wedge p_{\text{stop}} \ge 0.86 \wedge p_{\text{pos}} \ge 0.83)]$. While $t < B$, coarse views ($s_t > 4$) continue reading; at $s_t \le 4$, $\mathcal{E}_+(o_t)=1$ stops Positive, $H_t \ge 4 \wedge p_{\text{stop}}(o_t) \ge 0.50$ triggers an active Negative stop, and otherwise root switching selects $\arg\max_{c \in C_t^{\text{root,unvis}}} s_{\text{cand}}(c \mid o_t)$ (over $r_i=1$ roots at distance $\ge 0.05$ from visited $\text{DS}{=}64$ centers, with a full-pool argmax fallback) whenever the last two views have $s \le 4$ or $t \ge 3$ with $s_t \le 4$. At the safety horizon $t = B$, further reads are disabled and the prediction is Positive iff $\mathcal{E}_+(o_B) = 1$.
    \item Threshold-Calibrated Trace Readout and Budget Scaling: Both Stage-1 SFT and Stage-2 Traj-GRPO can be evaluated under either Online multi-region orchestration or a calibrated high-magnification threshold readout (Calib.: first crossing of $s_t \le 4, p_{\text{stop}}(o_t) \ge 0.50, p_{\text{pos}}(o_t) \ge 0.70$ over $B{=}12$ trajectories, matching Table~\ref{tab:camelyon_main}). Test-time budget scaling across $B \in \{4, 6, 8, 10, 12\}$ is evaluated via prefix truncation of $B{=}12$ trajectories (assigning Negative when an unfinished Stage-1 SFT trajectory is truncated at $t = B < 12$).
\end{enumerate}

\subsubsection{Query-Gated Multi-Scale Evidence Readout for WSI-VQA}
Alongside primary pure Step-$T$ evaluation ($\arg\max_j p_{\text{ans}}(a_j \mid o_T)$), we evaluate a query-gated multi-scale readout reusing Step-0 and Step-$T$ probabilities without additional model forwards:
\begin{equation}
    \label{eq:multiscale_readout}
    \hat{y} = \arg\max_{a_j \in \mathcal{A}} \!\Big[ (1 - \alpha_{m_Q})\,p_{\text{ans}}(a_j \mid o_0) + \alpha_{m_Q}\,p_{\text{ans}}(a_j \mid o_T) \Big],
\end{equation}
where $m_Q \in \{\text{Macro}, \text{Focal}\}$ denotes the macro-architectural versus focal-microscopic question category ($\alpha_{\text{Macro}} = 0.5$ to integrate global architectural context; $\alpha_{\text{Focal}} = 1.0$ to preserve pure Step-$T$ cellular readout).

\section{Experiments}
\label{sec:experiments}

We evaluate JevNavAgent across two complementary gigapixel whole-slide benchmarks covering multi-task visual question answering and lymph node metastasis screening and localization.

\subsection{Datasets, Task Stratification, and Evaluation Protocols}
\label{subsec:datasets_setup}

\begin{table}[t]
\centering
\footnotesize
\caption{Summary of Gigapixel WSI Benchmarks and Slide-Disjoint Splits. All tasks share the same action pyramid $\mathcal{S}=\{64, 16, 4, 1\}$.}
\label{tab:datasets}
\renewcommand{\arraystretch}{1.12}
\setlength{\tabcolsep}{2.2pt}
\begin{tabular}{llcc}
\toprule
\textbf{Benchmark} & \textbf{Split / Diagnostic Subset} & \textbf{WSIs} & \textbf{Cases} \\
\midrule
\multirow{6}{*}{\textbf{WSI-VQA}} 
 & Train Split (Slide-Disjoint) & $332$ & $1{,}002$ \\
 & Official Val / Test Splits & $77 / 76$ & $179 / 172$ \\
 \cmidrule(lr){2-4}
 & $\hookrightarrow$ \emph{Macro-Global Queries} & $142$ & $207$ \\
 & $\hookrightarrow$ \emph{Micro-Focal Queries} & $98$ & $144$ \\
 & $\hookrightarrow$ \emph{Buckets A--D: Standard Clin.} & $147$ & $296$ \\
 & $\hookrightarrow$ \emph{Bucket E: Multi-Finding Desc.} & $51$ & $55$ \\
\midrule
\multirow{3}{*}{\textbf{CAMELYON16}} 
 & Train Split ($68\text{ Pos} / 68\text{ Neg}$) & $136$ & $136$ \\
 & Dev Split ($23\text{ Pos} / 23\text{ Neg}$) & $46$ & $46$ \\
 & Official Test Split ($49\text{ Pos} / 80\text{ Neg}$) & $129$ & $129$ \\
\bottomrule
\end{tabular}\\[2pt]
{\scriptsize Note: Macro-Global ($207$) + Micro-Focal ($144$) = $351$ Val+Test queries across $153$ unique held-out WSIs; Buckets A--D ($296$) + Bucket E ($55$) = $351$ queries.}
\end{table}

Table~\ref{tab:datasets} summarizes the slide-disjoint splits and diagnostic query groupings of the two benchmarks. Crucially, across all splits and query types, the environment exposes the exact same multi-scale candidate pyramid $\mathcal{S} = \{64, 16, 4, 1\}$ ($0.625\times\text{--}40\times$); during active navigation, the policy receives no query-category metadata and must autonomously decide where to look, which magnification to select, and when to stop.

\subsubsection{Benchmark 1: Clean Visual WSI-VQA}
Existing pathology VQA datasets derived from pathology reports frequently contain non-visual clinical metadata leakage (e.g., patient age, surgical margin codes, or TNM staging numbers that cannot be inferred from histology alone) or text-only shortcuts solvable without image inspection. To rigorously evaluate active visual perception, we construct and evaluate on the Clean Visual WSI-VQA benchmark comprising $1{,}353$ model-audited multiple-choice questions across $485$ gigapixel TCGA-BRCA H\&E whole-slide images. All questions undergo model-assisted visual-grounding curation to filter non-histological staging leakage and are partitioned into slide-disjoint splits of $1{,}002$ training questions ($332$ WSIs), $179$ official validation questions ($77$ unique WSIs), and $172$ official test questions ($76$ unique WSIs), yielding $351$ held-out evaluation questions across $153$ unseen WSIs.

For fine-grained post-hoc diagnostic analysis, the $351$ evaluation questions are analyzed along two complementary partitions:
\begin{itemize}
    \item Diagnostic Scale Partition (Macro-Global vs.\ Micro-Focal): Macro-Global ($207$ questions across $142$ WSIs) covers architectural queries (e.g., predominant growth pattern, histological subtype, necrosis extent), whereas Micro-Focal ($144$ questions across $98$ WSIs) targets localized cellular phenomena (e.g., mitotic activity at $40\times$, nuclear atypia, focal lymphovascular or perineural invasion) that are imperceptible in the global thumbnail and require autonomous zooming into $\text{DS} \in \{4, 1\}$ fields.
    \item Clinical Task Partition (Buckets A--E): Bucket A ($96$ questions) assesses histological grading; Bucket B ($66$ questions) evaluates tumor subtyping; Bucket C ($45$ questions) examines invasion status; Bucket D ($89$ questions) tests focal presence and exclusion verification (including binary Yes/No and negative-exclusion microscopic findings); and Bucket E ($55$ questions) requires composite multi-finding synthesis across long descriptive statements.
\end{itemize}

\paragraph{WSI-VQA Evaluation Metrics}
Beyond overall accuracy on Val ($179$), Test ($172$), and Total ($351$), as well as accuracy on Macro-Global, Micro-Focal, and standard clinical Buckets A--D ($296$ questions), we report average read steps per episode, end-to-end wall-clock latency per episode (seconds), and three within-episode evidence integration diagnostics:
\begin{enumerate}
    \item Step-0 vs.\ Step-$T$ Accuracy: Diagnostic accuracy evaluated at the initial overview state $o_0$ (before acquiring any high-resolution crops) versus the terminal state $o_T$ (after active multi-scale navigation).
    \item Net Zoom Gain ($\Delta_{\text{zoom}}$): The net number of questions flipped from incorrect at Step 0 to correct at Step $T$ ($N_{0 \to 1}$) minus those flipped from correct to incorrect ($N_{1 \to 0}$), directly quantifying the evidentiary value of active zooming.
    \item Step-0/Step-$T$ Agreement and Belief Shift: The percentage of episodes where the terminal prediction remains identical to the Step-0 thumbnail prediction, paired with the mean $L_1$ probability shift $\|p_{\text{ans}}(\cdot \mid o_T) - p_{\text{ans}}(\cdot \mid o_0)\|_1$.
\end{enumerate}

\subsubsection{Benchmark 2: Exhaustively Verified CAMELYON16 Gigapixel Metastasis Screening and Localization}
While WSI-VQA evaluates guided diagnostic reasoning across breast pathology slides (TCGA-BRCA), sentinel lymph node metastasis detection in breast cancer represents a classic ``needle-in-a-haystack'' search problem spanning macrometastases and sub-millimeter micrometastases (often occupying $<0.5\%$ of the tissue area). We evaluate on the CAMELYON16 benchmark~\cite{bejnordi2017diagnostic} across raw gigapixel whole-slide images acquired from two independent medical centers: Radboud University Medical Center (Center 0) and University Medical Center Utrecht (Center 1).

\paragraph{Strict Exclusion of Partially Annotated Slides}
A critical yet frequently overlooked specification in the official CAMELYON16 documentation is that $20$ positive training slides carry incomplete, non-exhaustive polygon annotations, meaning unannotated tumor clusters exist outside the ground-truth XML masks. Using such slides to evaluate or supervise spatial localization introduces severe label noise, as an agent landing on an unannotated metastasis is falsely penalized as a benign tissue read. We strictly exclude all $20$ partially annotated slides, retaining all $91$ exhaustively annotated positive training slides, $91$ center-matched negative training slides, and the complete official test cohort of $129$ slides ($49$ positive, $80$ negative). In total, our benchmark encompasses $311$ verified gigapixel WSIs ($\sim 535\,\text{GiB}$ on disk) partitioned into:
\begin{itemize}
    \item Train Split ($136$ WSIs): $68$ exhaustively annotated positive WSIs and $68$ center-matched negative WSIs;
    \item Development Split ($46$ WSIs): $23$ exhaustively annotated positive WSIs and $23$ center-matched negative WSIs;
    \item Official Test Split ($129$ WSIs): All $49$ positive and $80$ negative official test WSIs.
\end{itemize}

\paragraph{CAMELYON16 Evaluation Metrics}
Unlike passive patch-level MIL classifiers that process thousands of pre-extracted tiles offline, an active visual navigation agent must actively locate metastases within a compact step budget $B \in \{4, \dots, 12\}$ while accurately ruling out negative slides. We evaluate four complementary metrics:
\begin{enumerate}
    \item Positive Localization Hit Rate (Pos Loc Hit): A strict spatial-and-diagnostic metric on positive slides requiring the agent to (i) terminate at cellular magnification ($s_T \in \{4, 1\}$), (ii) position its terminal crop center inside a ground-truth metastasis polygon with level-0 $512\times 512$ tumor area fraction $\ge 0.25$, and (iii) output a Positive diagnosis.
    \item Positive Visited Hit Rate (Visited Hit): Measures search trajectory coverage, defined as the fraction of positive slides on which the agent's camera visits a ground-truth tumor field satisfying the high-magnification spatial criterion ($s_t \in \{4, 1\}$) at least once during the episode. The ratio Pos Loc Hit\,/\,Visited Hit defines the stop conversion rate, i.e., the fraction of visited tumors that the agent converts into a grounded terminal stop (e.g., $11/12 = 91.7\%$ for Traj-GRPO on CAM Official Test).
    \item Negative Exclusion Specificity (Neg Spec): The fraction of negative slides correctly classified as Negative upon episode termination.
    \item Strict Loc+Spec and Overall Slide Accuracy (Slide Acc): Strict joint screening accuracy requiring both exact spatial tumor localization on positive slides and correct exclusion on negative slides, alongside overall binary slide-level classification accuracy.
\end{enumerate}

\subsubsection{Implementation and Training Protocol}
We initialize JevNavAgent from the open-source Qwen3-VL-2B-Instruct backbone~\cite{bai2025qwen3vl} ($2\,\text{B}$ parameters) and fine-tune the LoRA backbone weights jointly with the three factorized decision heads using AdamW ($\text{lr} = 2 \times 10^{-5}$, cosine decay, warmup ratio $0.05$, gradient accumulation $4$ / effective batch size $4$, $\text{bf16}$ precision). In accordance with our counterfactual trajectory alignment framework (Section~\ref{subsec:debiasing}), the Stage-1 training corpus $\mathcal{D}_{\text{train}}$ comprises $16{,}635$ state-action transitions ($15{,}875$ distinct trajectory states) across $449$ active training WSIs ($332$ WSI-VQA WSIs + $117$ CAMELYON16 WSIs with verified teacher rollouts from the $136$-WSI train split), where positive and negative terminal stops on the long-horizon CAMELYON16 subset ($t \in \{6, \dots, 12\}$) are balanced $1:1$ across each horizon $t$ and terminal scale $s_T \in \{4, 1\}$, alongside upsampled Center-1 negative $\text{READ}$ states and focal affirmative VQA stops. During Stage-1 SFT, overview dropout is applied with probability $0.5$ on terminal $\text{STOP}$ states ($t \ge 1$). Benefiting from single-forward execution, Stage-2 online Trajectory-Level GRPO runs on a single GPU in $1.04\,\text{h}$ for $30$-step VQA+KL ($\approx 0.7\,\text{h}$ to update 20) and $4.18\,\text{h}$ for $60$-step CAM16 ($\approx 3.1\,\text{h}$ to update 45). All latency measurements report the mean episode wall-clock seconds recorded in each method's evaluation logs.

\subsection{Main Results on Clean Visual WSI-VQA}
\label{subsec:exp_wsivqa}

\begin{table*}[t]
\centering
\footnotesize
\caption{Comparison on the 351-Question Clean Visual WSI-VQA Benchmark ($179$ Val + $172$ Test). Shaded columns (Total$^\star$ and Micro-Focal$^\star$) denote primary evaluation metrics ($\%$). Green superscripts indicate gains over Qwen3-VL-8B + PLIP.}
\label{tab:wsivqa_main}
\renewcommand{\arraystretch}{1.16}
\setlength{\tabcolsep}{1.4pt}
\begin{tabular}{llccc>{\columncolor{priCol}}cc>{\columncolor{priCol}}cccc}
\toprule
\textbf{Category} & \textbf{Method} & \textbf{Params} & \textbf{Val} & \textbf{Test} & \textbf{Total$^\star$} & \textbf{Macro-Global} & \textbf{Micro-Focal$^\star$} & \textbf{Clin.\ A--D} & \textbf{Reads} & \textbf{Time (s)} \\
\midrule
\multirow{2}{*}{\emph{I.\ Frontier Agents}}
 & GPT-6.1-Terra & Frontier & $41.90$ & $51.16$ & $46.44$ & $52.17$ & $38.19$ & $45.61$ & $3.33$ & $70.04$ \\
 & GPT-6.1-Sol & Frontier & $48.60$ & $51.16$ & $49.86$ & $53.14$ & $45.14$ & $50.00$ & $2.69$ & $66.64$ \\
\midrule
\multirow{3}{*}{\emph{II.\ Open VLMs (ZS)}}
 & Patho-R1-7B & $7\text{B}$ & $32.96$ & $38.95$ & $35.90$ & $36.71$ & $34.72$ & $33.78$ & $0.00$ & $0.78$ \\
 & Quilt-LLaVA-7B & $7\text{B}$ & $37.99$ & $39.53$ & $38.75$ & $34.78$ & $44.44$ & $36.15$ & $0.00$ & $0.68$ \\
 & Qwen3-VL-8B & $8\text{B}$ & $40.22$ & $45.35$ & $42.74$ & $48.79$ & $34.03$ & $41.22$ & $0.00$ & $0.66$ \\
\midrule
\multirow{6}{*}{\emph{III.\ Competing Nav}}
 & Quilt-LLaVA + TissueNav & $7\text{B}$ & $40.22$ & $36.63$ & $38.46$ & $35.27$ & $43.06$ & $36.49$ & $3.00$ & $1.54$ \\
 & PathNavigate & $7\text{B}$ & $36.31$ & $43.02$ & $39.60$ & $38.65$ & $40.97$ & $38.51$ & $2.98$ & $1.94$ \\
 & PathAgent & $11\text{B}$ & $37.99$ & $42.44$ & $40.17$ & $47.34$ & $29.86$ & $39.19$ & $2.98$ & $3.96$ \\
 & Qwen3-VL-2B + PLIP & $2\text{B}$ & $39.66$ & $50.58$ & $45.01$ & $51.21$ & $36.11$ & $45.95$ & $2.98$ & $1.83$ \\
 & Qwen3-VL-8B + PLIP & $8\text{B}$ & $44.13$ & $49.42$ & $46.72$ & $51.69$ & $39.58$ & $44.93$ & $2.98$ & $1.88$ \\
 & MedGemma-4B + PLIP & $4\text{B}$ & $50.28$ & $50.58$ & $50.43$ & $\mathbf{53.14}$ & $46.53$ & $51.35$ & $2.98$ & $2.53$ \\
\midrule
\multirow{5}{*}{\emph{IV.\ 2B Nav Controls}}
 & Thumbnail-Only (Step 0) & $2\text{B}$ & $42.46$ & $37.21$ & $39.89$ & $43.96$ & $34.03$ & $41.89$ & $0.00$ & $\mathbf{0.44}$ \\
 & Random-Nav (2B) & $2\text{B}$ & $48.04$ & $48.84$ & $48.43$ & $47.83$ & $49.31$ & $51.69$ & $3.00$ & $1.57$ \\
 & TissuePeak-Nav (2B) & $2\text{B}$ & $47.49$ & $50.00$ & $48.72$ & $49.28$ & $47.92$ & $53.38$ & $3.00$ & $1.68$ \\
 & SurpriseNav (2B) & $2\text{B}$ & $48.04$ & $49.42$ & $48.72$ & $50.24$ & $46.53$ & $54.05$ & $2.98$ & $1.64$ \\
 & PLIP-Nav (2B) & $2\text{B}$ & $48.04$ & $50.00$ & $49.00$ & $49.28$ & $48.61$ & $53.38$ & $2.98$ & $1.78$ \\
\midrule
 & JevNavAgent-M1 (SFT) & $2\text{B}$ & $46.93$ & $50.00$ & $48.43$ & $49.76$ & $46.53$ & $53.38$ & $3.11$ & $4.78$ \\
 & JevNavAgent-M2 (SFT) & $2\text{B}$ & $45.25$ & $50.58$ & $47.86$ & $47.34$ & $48.61$ & $52.36$ & $3.13$ & $4.91$ \\
 & JevNavAgent-M3 (SFT) & $2\text{B}$ & $43.58$ & $50.00$ & $46.72$ & $46.86$ & $46.53$ & $51.35$ & $3.11$ & $6.21$ \\
 & JevNavAgent (SFT, Step-$T$) & $2\text{B}$ & $50.84$ & $48.84$ & $49.86$ \gain{3.14} & $50.24$ & $49.31$ \gain{9.73} & $53.38$ & $3.09$ & $5.25$ \\
\rowcolor{oursRow}
\emph{V.\ Ours} & \textbf{JevNavAgent + Traj-GRPO (Step-$T$)} & $2\text{B}$ & $50.84$ & $50.00$ & $50.43$ \gain{3.71} & $49.76$ & $\mathbf{51.39}$ \gain{11.81} & $53.38$ & $3.05$ & $5.63$ \\
\rowcolor{oursRow}
\emph{(2B)} & \textbf{JevNavAgent + Traj-GRPO (MS)} & $2\text{B}$ & $\mathbf{51.40}$ & $\mathbf{51.74}$ & $\mathbf{51.57}$ \gain{4.85} & $51.69$ & $\mathbf{51.39}$ \gain{11.81} & $\mathbf{54.73}$ & $3.05$ & $5.63$ \\
\bottomrule
\end{tabular}\\[3pt]
{\scriptsize Note: $^\star$Primary metrics: Total$^\star$ (overall accuracy across all $351$ queries) and Micro-Focal$^\star$ ($144$ cellular queries requiring $10\times\text{--}40\times$ inspection). \gain{Gain} is relative to $8\text{B}$ Qwen3-VL-8B + PLIP. PathNavigate and PathAgent are adapted under our unified quadtree candidate pool and fixed step budget for matched-environment comparison. Category V wall-clock latency is recorded under eager-attention evaluation ($4.78\text{--}6.21\,\text{s}$).}
\end{table*}

Table~\ref{tab:wsivqa_main} presents the comparison across five model categories on the 351-question Clean Visual WSI-VQA benchmark ($179$ Official Val + $172$ Official Test). We highlight two primary findings:

\subsubsection{Comparison with Frontier Agents and 4B--11B Open-Source Systems}
Despite using a compact 2B-parameter backbone, Stage-1 JevNavAgent (SFT, Step-$T$) reaches $49.86\%$ overall accuracy ($175/351$). Stage-2 online optimization (JevNavAgent + Traj-GRPO) lifts pure Step-$T$ accuracy to $50.43\%$ ($177/351$, requiring zero query-type metadata), matching the strongest 4B cascade (MedGemma-4B + PLIP) overall while outperforming it on Micro-Focal queries ($51.39\%$ vs.\ $46.53\%$). Under query-gated multi-scale readout (Eq.~\eqref{eq:multiscale_readout}), JevNavAgent + Traj-GRPO reaches $51.57\%$ ($181/351$; $51.40\%$ Val, $51.74\%$ Test), surpassing frontier API agents, 7B--8B zero-shot VLMs, and adapted multi-model navigation pipelines across all $351$ queries. Two failure modes explain the gap among larger baselines. First, when restricted to static thumbnails ($\text{DS}=64$), larger instruction-tuned VLMs exhibit affirmative hallucination bias on unresolved microscopic features, over-predicting ``Yes/Present'' ($23.60\%$ for Qwen3-VL-8B and $26.97\%$ for Patho-R1-7B on Bucket D), whereas pairing them with high-magnification navigation restores positive scaling ($46.72\%$ vs.\ $45.01\%$). Second, text-mediated cascades such as PathAgent lose fine cellular detail across intermediate summaries ($29.86\%$ Micro-Focal), whereas JevNavAgent performs end-to-end latent readout in $5.25\text{--}5.63\,\text{s}$ ($11.8\times\text{--}12.4\times$ faster than frontier API agents).

\subsubsection{Matched-Backbone Control Verification and 5-Bucket Clinical Breakdown (Table~\ref{tab:wsivqa_buckets})}
Under strict matched-backbone controls (Category IV, fixing the same 2B answer head while varying only the search policy), simple 2B navigators from Random-Nav to PLIP-Nav already recover most of the gap over static thumbnails ($39.89\% \to 48.43\%\text{--}49.00\%$). This reveals where learned navigation matters most: rather than broad architectural cues that almost any high-resolution crop exposes, the additional margin of JevNavAgent concentrates on cellular-level Micro-Focal and Bucket D queries. Specifically, rewarding step-wise confidence gain $\Delta P_t$ via Traj-GRPO lifts Micro-Focal accuracy ($144$ instances requiring $10\times\text{--}40\times$ inspection) from $49.31\%$ to $51.39\%$, outperforming Qwen3-VL-8B + PLIP by $+11.81\text{ pp}$, GPT-6.1-Sol by $+6.25\text{ pp}$, and MedGemma-4B + PLIP by $+4.86\text{ pp}$.

\begin{table}[t]
\centering
\footnotesize
\caption{Fine-Grained 5-Bucket Clinical Breakdown on Clean Visual WSI-VQA ($351$ Queries, $\%$): A Grade ($96$), B Subtype ($66$), C Invasion ($45$), D$^\star$ Focal Pres./Excl.\ ($89$), E Multi-Finding Desc.\ ($55$), and A--D$^\star$ ($296$).}
\label{tab:wsivqa_buckets}
\renewcommand{\arraystretch}{1.15}
\setlength{\tabcolsep}{0.7pt}
\begin{tabular}{lccc>{\columncolor{priCol}}cc>{\columncolor{priCol}}c}
\toprule
\textbf{Method} & \textbf{A} & \textbf{B} & \textbf{C} & \textbf{D$^\star$} & \textbf{E} & \textbf{A--D$^\star$} \\
\midrule
Patho-R1-7B (ZS)      & $16.67$ & $63.64$ & $40.00$ & $26.97$ & $47.27$ & $33.78$ \\
Quilt-LLaVA-7B (ZS)   & $21.88$ & $46.97$ & $44.44$ & $39.33$ & $52.73$ & $36.15$ \\
Qwen3-VL-8B (ZS)      & $37.50$ & $62.12$ & $\mathbf{53.33}$ & $23.60$ & $50.91$ & $41.22$ \\
PathNavigate (7B)     & $19.79$ & $63.64$ & $42.22$ & $38.20$ & $45.45$ & $38.51$ \\
PathAgent (11B)       & $36.46$ & $\mathbf{69.70}$ & $37.78$ & $20.22$ & $45.45$ & $39.19$ \\
Qwen3-VL-8B + PLIP    & $44.79$ & $62.12$ & $51.11$ & $29.21$ & $\mathbf{56.36}$ & $44.93$ \\
MedGemma-4B + PLIP    & $48.96$ & $63.64$ & $46.67$ & $47.19$ & $45.45$ & $51.35$ \\
GPT-6.1-Terra         & $47.92$ & $65.15$ & $42.22$ & $30.34$ & $50.91$ & $45.61$ \\
GPT-6.1-Sol           & $\mathbf{51.04}$ & $63.64$ & $42.22$ & $42.70$ & $49.09$ & $50.00$ \\
\midrule
JevNavAgent (SFT)     & $41.67$ & $62.12$ & $51.11$ & $60.67$ & $30.91$ & $53.38$ \\
\rowcolor{oursRow}
\textbf{+ GRPO (Step-$T$)} & $43.75$ & $62.12$ & $44.44$ & $\mathbf{61.80}$ \gain{32.6} & $34.55$ & $53.38$ \gain{8.5} \\
\rowcolor{oursRow}
\textbf{+ GRPO (MS)} & $42.71$ & $66.67$ & $48.89$ & $\mathbf{61.80}$ \gain{32.6} & $34.55$ & $\mathbf{54.73}$ \gain{9.8} \\
\bottomrule
\end{tabular}\\[2pt]
{\scriptsize Note: On Bucket D$^\star$ ($89$ focal presence and exclusion queries), PathAgent, Qwen3-VL-8B, Patho-R1-7B, and Qwen3-VL-8B + PLIP score $20.22\%\text{--}29.21\%$, whereas JevNavAgent + GRPO reaches $61.80\%$ (\gain{Gains} vs.\ 8B + PLIP).}
\end{table}

Table~\ref{tab:wsivqa_buckets} decomposes performance across the five clinical diagnostic buckets: (A) Histologic Grade ($96$), (B) Tumor Subtype \& Architecture ($66$), (C) Margin \& Lymphovascular Invasion ($45$), (D) Focal Presence \& Exclusion Verification ($89$), and (E) Multi-Finding Description ($55$). The contrast between Buckets D and E clarifies the division between active visual perception and raw language scale. On Bucket D, where binary Yes/No and negative-exclusion findings depend critically on active cellular inspection, all open and frontier baselines remain below $48\%$ ($20.22\%\text{--}47.19\%$), whereas JevNavAgent + Traj-GRPO reaches $61.80\%$ ($55/89$, $+32.59\text{ pp}$ over Qwen3-VL-8B + PLIP and $+19.10\text{ pp}$ over GPT-6.1-Sol), driving $54.73\%$ ($162/296$) across standard clinical Buckets A--D. Conversely, on Bucket E ($34.55\%$ vs.\ $45.45\%\text{--}56.36\%$ for 7B--8B models), multi-finding descriptive synthesis across long composite statements benefits more from broad LLM language capacity than from single-focus high-magnification verification.

\subsection{Gigapixel Metastasis Localization, Negative Exclusion, and Test-Time Budget Scaling}
\label{subsec:exp_camelyon}

\begin{table*}[t]
\centering
\footnotesize
\caption{Comparison on CAMELYON16 ($46$ Dev WSIs and $129$ Official Test WSIs) Under a Unified Inference Budget ($B = 12$). Shaded columns (Pos Loc Hit$^\star$, Strict Loc+Spec$^\star$, and Slide Acc$^\star$) denote the primary localization, joint grounded-screening, and slide-level diagnostic metrics ($\%$).}
\label{tab:camelyon_main}
\renewcommand{\arraystretch}{1.15}
\setlength{\tabcolsep}{1.1pt}
\begin{tabular}{llc>{\columncolor{priCol}}cccc>{\columncolor{priCol}}c>{\columncolor{priCol}}cc}
\toprule
\textbf{Split} & \textbf{Method} & \textbf{Stop Mode} & \textbf{Pos Loc Hit$^\star$} & \textbf{Visited Hit} & \textbf{Pos Recall} & \textbf{Neg Spec} & \textbf{Strict Loc+Spec$^\star$} & \textbf{Slide Acc$^\star$} & \textbf{Reads (Pos/Neg)} \\
\midrule
\multirow{10}{*}{\textbf{Dev ($46$)}}
 & GPT-6.1-Terra (Frontier) & Dynamic & $17.39$ & $17.39$ & $95.65$ & $52.17$ & $34.78$ & $\mathbf{73.91}$ & $3.83 / 4.30$ \\
 & GPT-6.1-Sol (Frontier) & Dynamic & $\mathbf{30.43}$ & $\mathbf{30.43}$ & $95.65$ & $52.17$ & $41.30$ & $\mathbf{73.91}$ & $4.43 / 3.57$ \\
 \cmidrule(lr){2-10}
 & PathNavigate (7B) & Fixed-$B$ & $0.00$ & $21.74$ & $100.0$ & $0.00$ & $0.00$ & $50.00$ & $12.00 / 12.00$ \\
 & PathAgent (11B) & Fixed-$B$ & $8.70$ & $\mathbf{30.43}$ & $100.0$ & $0.00$ & $4.35$ & $50.00$ & $12.00 / 12.00$ \\
 & MedGemma-4B + PLIP & Fixed-$B$ & $8.70$ & $\mathbf{30.43}$ & $95.65$ & $13.04$ & $10.87$ & $54.35$ & $12.00 / 12.00$ \\
 & Qwen3-VL-8B + PLIP & Fixed-$B$ & $8.70$ & $\mathbf{30.43}$ & $100.0$ & $0.00$ & $4.35$ & $50.00$ & $12.00 / 12.00$ \\
 \cmidrule(lr){2-10}
 & Random-Nav (2B) & Fixed-$B$ & $0.00$ & $0.00$ & $0.00$ & $91.30$ & $45.65$ & $45.65$ & $12.00 / 12.00$ \\
 & Grid-Nav (2B) & Fixed-$B$ & $4.35$ & $4.35$ & $39.13$ & $52.17$ & $28.26$ & $45.65$ & $12.00 / 12.00$ \\
 & SurpriseNav (2B) & Fixed-$B$ & $0.00$ & $21.74$ & $8.70$ & $73.91$ & $36.96$ & $41.30$ & $12.00 / 12.00$ \\
 & PLIP-Nav (2B) & Fixed-$B$ & $8.70$ & $\mathbf{30.43}$ & $17.39$ & $86.96$ & $47.83$ & $52.17$ & $12.00 / 12.00$ \\
\rowcolor{oursRow}
 & \textbf{JevNavAgent (SFT, 2B)} & \textbf{Dynamic} & $\mathbf{30.43}$ \gain{21.7} & $\mathbf{30.43}$ & $47.83$ & $\mathbf{95.65}$ & $\mathbf{63.04}$ \gain{15.2} & $71.74$ \gain{17.4} & $\mathbf{9.91 / 11.91}$ \\
\rowcolor{oursRow}
\textbf{Dev} & \textbf{JevNavAgent + Traj-GRPO (2B)} & \textbf{Dynamic} & $21.74$ \gain{13.0} & $26.09$ & $30.43$ & $91.30$ & $56.52$ \gain{8.7} & $60.87$ \gain{6.5} & $10.70 / 11.91$ \\
\midrule
\multirow{10}{*}{\textbf{Test ($129$)}}
 & GPT-6.1-Terra (Frontier) & Dynamic & $18.37$ & $18.37$ & $73.47$ & $27.50$ & $24.03$ & $44.96$ & $3.67 / 3.80$ \\
 & GPT-6.1-Sol (Frontier) & Dynamic & $10.20$ & $10.20$ & $51.02$ & $48.75$ & $34.11$ & $49.61$ & $3.45 / 3.61$ \\
 \cmidrule(lr){2-10}
 & PathNavigate (7B) & Fixed-$B$ & $0.00$ & $22.45$ & $100.0$ & $0.00$ & $0.00$ & $37.98$ & $12.00 / 12.00$ \\
 & PathAgent (11B) & Fixed-$B$ & $12.24$ & $\mathbf{26.53}$ & $100.0$ & $0.00$ & $4.65$ & $37.98$ & $12.00 / 12.00$ \\
 & MedGemma-4B + PLIP & Fixed-$B$ & $12.24$ & $\mathbf{26.53}$ & $89.80$ & $15.00$ & $13.95$ & $43.41$ & $12.00 / 12.00$ \\
 & Qwen3-VL-8B + PLIP & Fixed-$B$ & $12.24$ & $\mathbf{26.53}$ & $100.0$ & $1.25$ & $5.43$ & $38.76$ & $12.00 / 12.00$ \\
 \cmidrule(lr){2-10}
 & Random-Nav (2B) & Fixed-$B$ & $0.00$ & $4.08$ & $4.08$ & $95.00$ & $58.91$ & $60.47$ & $12.00 / 12.00$ \\
 & Grid-Nav (2B) & Fixed-$B$ & $4.08$ & $4.08$ & $53.06$ & $51.25$ & $33.33$ & $51.94$ & $12.00 / 12.00$ \\
 & SurpriseNav (2B) & Fixed-$B$ & $0.00$ & $22.45$ & $18.37$ & $86.25$ & $53.49$ & $60.47$ & $12.00 / 12.00$ \\
 & PLIP-Nav (2B) & Fixed-$B$ & $6.12$ & $\mathbf{26.53}$ & $14.29$ & $88.75$ & $57.36$ & $60.47$ & $12.00 / 12.00$ \\
 \cmidrule(lr){2-10}
 & JevNavAgent (SFT, Online) & Dynamic & $10.20$ & $22.45$ & $30.61$ & $88.75$ & $58.91$ & $66.67$ \gain{6.2} & $10.16 / 11.45$ \\
 & JevNavAgent (SFT, Calib.) & Dynamic & $14.29$ & $14.29$ & $79.59$ & $23.75$ & $20.16$ & $44.96$ & $7.29 / 7.39$ \\
\rowcolor{oursRow}
 & \textbf{JevNavAgent + Traj-GRPO (Online)} & \textbf{Dynamic} & $16.33$ \gain{4.1} & $24.49$ & $20.41$ & $\mathbf{98.75}$ & $\mathbf{67.44}$ \gain{8.5} & $68.99$ \gain{8.5} & $10.96 / 11.97$ \\
\rowcolor{oursRow}
\textbf{Test} & \textbf{JevNavAgent + Traj-GRPO (Calib.)} & \textbf{Dynamic} & $\mathbf{22.45}$ \gain{10.2} & $24.49$ & $30.61$ & $93.75$ & $66.67$ \gain{7.8} & $\mathbf{69.77}$ \gain{9.3} & $\mathbf{10.29 / 11.75}$ \\
\bottomrule
\end{tabular}\\[3pt]
{\scriptsize Note: All methods are evaluated under $B=12$. Stop Mode: Fixed-$B$ retrieval/heuristic navigators (including our unified-environment adaptations of PathNavigate and PathAgent) exhaust all $12$ reads ($12.00/12.00$), whereas Dynamic agents decide at each step whether to continue reading or stop. On Test ($129$), we report both Stage-1 SFT and Stage-2 Traj-GRPO (CAM16-specialized) under matched stopping rules (Section~\ref{subsubsec:screening_orchestration}): Online multi-region orchestration (Online) and calibrated high-magnification threshold ($s_t \le 4, p_{\text{stop}} \ge 0.50, p_{\text{pos}} \ge 0.70$; Calib.). \gain{Gains} are over the strongest open-source/2B baseline (an always-Negative baseline on Test scores $0.00\%$ Loc Hit and $62.02\%$ Slide Acc / Strict Loc+Spec).}
\end{table*}

\begin{table}[t]
\centering
\footnotesize
\caption{Zero-Shot Test-Time Budget Scaling of JevNavAgent (SFT, 2B) Across $B \in \{4, 6, 8, 10, 12\}$ on CAMELYON16 ($\%$). Shaded columns (Loc Hit$^\star$ and Slide Acc$^\star$) denote primary metrics.}
\label{tab:budget_scaling}
\renewcommand{\arraystretch}{1.15}
\setlength{\tabcolsep}{1.7pt}
\begin{tabular}{lc>{\columncolor{priCol}}cccc>{\columncolor{priCol}}c}
\toprule
\textbf{Split} & \textbf{Budget} & \textbf{Loc Hit$^\star$} & \textbf{Visited} & \textbf{Pos Rec} & \textbf{Neg Spec} & \textbf{Slide Acc$^\star$} \\
\midrule
\multirow{4}{*}{\textbf{Dev ($46$)}}
 & $B = 4$  & $8.70$  & $8.70$  & $8.70$  & $\mathbf{100.0}$ & $54.35$ \\
 & $B = 6$  & $13.04$ & $17.39$ & $17.39$ & $\mathbf{100.0}$ & $58.70$ \\
 & $B = 8$  & $13.04$ & $17.39$ & $17.39$ & $\mathbf{100.0}$ & $58.70$ \\
 & $B = 10$ & $26.09$ & $\mathbf{30.43}$ & $43.48$ & $95.65$ & $69.57$ \\
\rowcolor{oursRow}
\textbf{Dev} & $\mathbf{B = 12}$ & $\mathbf{30.43}$ \gain{21.7} & $\mathbf{30.43}$ & $\mathbf{47.83}$ & $95.65$ & $\mathbf{71.74}$ \gain{17.4} \\
\midrule
\multirow{4}{*}{\textbf{Test ($129$)}}
 & $B = 4$  & $4.08$  & $10.20$ & $10.20$ & $\mathbf{97.50}$ & $64.34$ \\
 & $B = 6$  & $8.16$  & $16.33$ & $20.41$ & $93.75$ & $65.89$ \\
 & $B = 8$  & $8.16$  & $16.33$ & $20.41$ & $93.75$ & $65.89$ \\
 & $B = 10$ & $\mathbf{10.20}$ & $20.41$ & $\mathbf{30.61}$ & $90.00$ & $\mathbf{67.44}$ \\
\rowcolor{oursRow}
\textbf{Test} & $\mathbf{B = 12}$ & $\mathbf{10.20}$ \gain{6.1} & $\mathbf{22.45}$ & $\mathbf{30.61}$ & $88.75$ & $66.67$ \gain{2.3} \\
\bottomrule
\end{tabular}
\end{table}

Table~\ref{tab:camelyon_main} and Table~\ref{tab:budget_scaling} report spatial localization, negative exclusion, and test-time budget scaling performance on CAMELYON16 across the Development split ($46$ WSIs: $23$ Pos + $23$ Neg) and the Official Test split ($129$ WSIs: $49$ Pos + $80$ Neg).

\subsubsection{Active Multi-Branch Search and Trajectory-Level GRPO vs.\ Frontier and Fixed-Budget Agents}
Behavioral inspection of baseline rollouts on CAMELYON16 exposes two complementary failure modes. First, frontier closed-source VLM agents (GPT-6.1-Terra and GPT-6.1-Sol) commit irreversibly to their initial root selection and terminate after only $3.45\text{--}4.43$ reads without cross-region backtracking; on normal lymph nodes this causes severe false-positive hallucinations ($27.50\%\text{--}52.17\%$ specificity), limiting Official Test Strict Loc+Spec to $24.03\%\text{--}34.11\%$ and Slide Acc to $44.96\%\text{--}49.61\%$. Second, fixed-budget open-source pipelines (PathNavigate, PathAgent, MedGemma-4B + PLIP, Qwen3-VL-8B + PLIP) achieve solid trajectory coverage ($21.74\%\text{--}30.43\%$ Visited Hit under $B=12$) but lack counterfactual stopping alignment, over-calling benign lymphoid follicles as positive ($0.00\%\text{--}15.00\%$ specificity).

Table~\ref{tab:camelyon_main} shows that resolving screening accuracy hinges on the coupling between trajectory coverage and stop conversion. On Dev ($46$), Stage-1 JevNavAgent (SFT) matches GPT-6.1-Sol's top $30.43\%$ localization hit rate ($7/23$) at $95.65\%$ specificity ($63.04\%$ Strict Loc+Spec, $+21.74\text{ pp}$ over GPT-6.1-Sol). On Official Test ($129$, spanning $27$ micro- and $22$ macrometastases), Stage-1 SFT reaches $22.45\%$ Visited Hit ($11/49$) but converts only $5/11$ visited tumors into grounded terminal stops ($10.20\%$ Loc Hit) because state-wise imitation occasionally reads past a tumor focus. Stage-2 Traj-GRPO (Section~\ref{subsec:traj_grpo}) directly targets this bottleneck, lifting in-polygon stopping probability $p_{\text{stop}}(o_t \mid \text{Tumor})$ from $70.1\%$ (updates 1--10) to $94.9\%$ (updates 41--50, where update 45 is selected, before settling at $67.5\%$ in updates 51--60) while cutting benign-follicle false-positive risk from $7.58\%$ to $2.35\%$ (updates 31--40) and $3.69\%$ (updates 51--60). As a result, Stage-2 outperforms Stage-1 under both matched stopping protocols on Official Test. Under Online orchestration, Stage-2 converts visited tumors into grounded stops far more reliably ($5/11 \to 8/12$, $16.33\%$ Loc Hit) while holding $98.75\%$ specificity, achieving the highest Strict Loc+Spec of $67.44\%$ ($68.99\%$ Slide Acc). Under the simpler Calib.\ threshold, RL's benign-follicle suppression preserves $93.75\%$ specificity whereas SFT collapses to $23.75\%$; this enables Stage-2 to convert $11/12$ ($91.7\%$) visited tumors into grounded stops, reaching $22.45\%$ Loc Hit ($50.00\%$ on macrometastases, $11/22$, vs.\ $22.73\%$ for SFT), $66.67\%$ Strict Loc+Spec ($+32.56\text{ pp}$ over GPT-6.1-Sol), and $69.77\%$ Slide Acc (well above the $62.02\%$ always-Negative baseline). On the smaller $46$-slide Dev split ($23$ positives), Stage-2 shows modest split variance ($21.74\%$ vs.\ $30.43\%$ Loc Hit, a $2$-slide difference) while generalizing strongly to the larger $129$-slide Test set. Locating $<2\,\text{mm}$ micrometastases ($27$ Test slides, $<0.01\%$ slide area) within $12$ patches remains the primary physical bottleneck ($0.00\%$ localization).

\begin{figure*}[t]
\centering
\includegraphics[width=0.98\textwidth]{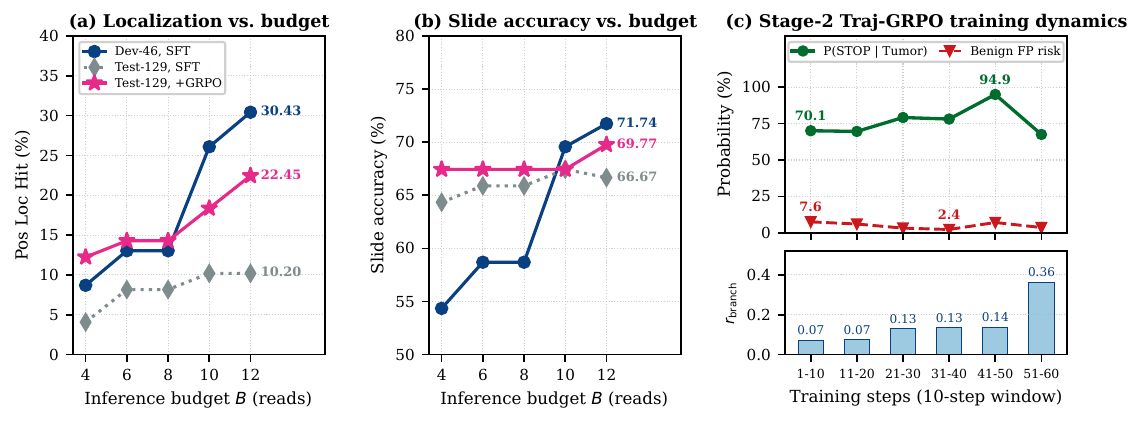}
\caption{Zero-shot test-time compute scaling and Stage-2 online reinforcement learning dynamics on CAMELYON16. \textbf{(a, b)} Decoupling the execution budget $B$ from the policy prompt lets the same checkpoint scale across $B \in \{4, 6, 8, 10, 12\}$ without retraining: Dev positive localization hit rate rises from $8.70\%$ to $30.43\%$ and slide accuracy from $54.35\%$ to $71.74\%$ (Stage-1 SFT); on the held-out Test split, Stage-2 Traj-GRPO raises the hit rate from $12.24\%$ to $22.45\%$ and slide accuracy from $67.44\%$ to $69.77\%$, while Stage-1 SFT reaches $10.20\%$ and $66.67\%$ (Traj-GRPO uses the calibrated stopping rule of Table~\ref{tab:camelyon_main}). \textbf{(c)} Across the 60-step Stage-2 training log (10-step window averages), in-polygon stopping probability $P(\text{STOP}\mid\text{Tumor})$ rises from $70.1\%$ (steps 1--10) to a peak of $94.9\%$ (steps 41--50) before settling at $67.5\%$ (steps 51--60), benign-follicle false-positive risk falls from $7.58\%$ to $2.35\%$ (steps 31--40), and the branch search reward increases $5.0\times$ ($0.072 \to 0.362$).}
\label{fig:scaling_grpo}
\end{figure*}

\subsubsection{Asymmetric Stopping Behavior and Zero-Shot Test-Time Budget Scaling (Table~\ref{tab:budget_scaling} and Fig.~\ref{fig:scaling_grpo})}
Because JevNavAgent decouples the step budget $B$ from the policy prompt (Section~\ref{subsec:pomdp}), a single checkpoint scales zero-shot across $B=4$ to $B=12$ without retraining (Table~\ref{tab:budget_scaling} and Fig.~\ref{fig:scaling_grpo}a,b). On Dev ($46$), expanding the horizon allows the agent to backtrack to second and third tissue roots after ruling out initial benign follicles, lifting Loc Hit from $8.70\%$ to $30.43\%$ ($+21.74\text{ pp}$) and Slide Acc from $54.35\%$ to $71.74\%$ while retaining $95.65\%$ specificity. On Official Test ($129$), scaling from $B=4$ to $B=10$ monotonically expands trajectory coverage ($10.20\% \to 20.41\%$ Visited Hit) and lifts Slide Acc from $64.34\%$ to $67.44\%$. Notably, Test exhibits a $B=10 \to 12$ inflection point: Visited Hit continues to grow ($20.41\% \to 22.45\%$), but allocating two additional reads slightly reduces specificity ($90.00\% \to 88.75\%$), causing SFT Test Slide Acc to peak at $B=10$ ($67.44\%$) before edging to $66.67\%$ at $B=12$.

\subsection{Diagnostic and Ablation Analysis: Uncovering and Breaking Non-Visual Shortcuts}
\label{subsec:exp_ablations}

\begin{table}[t]
\centering
\footnotesize
\caption{Within-Episode Zoom Diagnostics on WSI-VQA ($351$ Cases). Shaded columns (Step-$T^\star$ and $\Delta_{\text{zoom}}^\star$) highlight active navigation gains.}
\label{tab:zoom_diagnostics}
\renewcommand{\arraystretch}{1.16}
\setlength{\tabcolsep}{1.8pt}
\begin{tabular}{lc>{\columncolor{priCol}}c>{\columncolor{priCol}}ccc}
\toprule
\textbf{Ablation Variant} & \textbf{Step-0} & \textbf{Step-$T^\star$} & \textbf{$\Delta_{\text{zoom}}^\star$} & \textbf{Agree.} & \textbf{$L_1$ Shift} \\
\midrule
M0 (Shallow Head)   & $46.44$ & $47.01$ & $+2$  & $98.58$ & $0.021$ \\
M1 (+Gated CtxHead) & $46.15$ & $48.43$ & $+8$  & $64.39$ & $0.375$ \\
M2 (+Scale/Recency) & $50.14$ & $47.86$ & $-8$  & $76.35$ & $0.318$ \\
M3 (+Ov.\ Dropout)  & $48.43$ & $46.72$ & $-6$  & $\mathbf{58.69}$ & $\mathbf{0.424}$ \\
\rowcolor{oursRow}
\textbf{JevNavAgent (v5)} & $39.89$ & $49.86$ & $\mathbf{+35}$ \gain{9.97} & $\mathbf{58.69}$ & $0.396$ \\
\rowcolor{oursRow}
\textbf{+ Traj-GRPO} & $42.74$ & $\mathbf{50.43}$ & $+27$ \gain{7.69} & $65.81$ & $0.343$ \\
\bottomrule
\end{tabular}\\[2pt]
{\scriptsize Note: $\Delta_{\text{zoom}} = N_{0\to 1} - N_{1\to 0}$ (net flips right at Step $T$; green: accuracy gain in pp).}
\end{table}

\begin{table}[t]
\centering
\footnotesize
\caption{Ablation of Trajectory Distillation Shortcuts on CAMELYON16 Dev ($46$ WSIs, $B=12$).}
\label{tab:shortcut_ablation}
\renewcommand{\arraystretch}{1.16}
\setlength{\tabcolsep}{1.5pt}
\begin{tabular}{lcc>{\columncolor{priCol}}cc>{\columncolor{priCol}}c}
\toprule
\textbf{Regime} & \textbf{Pan $(16,4,4)$} & \textbf{Reads (P/N)} & \textbf{Loc Hit$^\star$} & \textbf{Spec} & \textbf{Acc$^\star$} \\
\midrule
1.\ Pos-Only   & $66\text{:}0$  & $3.57/3.00$  & $21.74$ & $0.00$  & $50.00$ \\
2.\ Unaligned  & $0\text{:}70$  & $3.83/8.17$  & $13.04$ & $30.43$ & $65.22$ \\
3.\ READ-Neg   & $12\text{:}13$ & $11.78/12.00$ & $0.00$  & $\mathbf{100.0}$ & $50.00$ \\
\rowcolor{oursRow}
4.\ \textbf{Ours (v5)} & $\mathbf{12\text{:}13}$ & $\mathbf{9.91/11.91}$ & $\mathbf{30.43}$ \gain{8.7} & $95.65$ & $\mathbf{71.74}$ \gain{6.5} \\
\bottomrule
\end{tabular}\\[2pt]
{\scriptsize Note: Pan $(16,4,4)$ is Pos:Neg stop count after lateral panning in each training corpus; Reads (P/N) denotes mean reads on positive/negative slides; \gain{Loc Hit} is vs.\ Row 1 and \gain{Acc} is vs.\ Row 2.}
\end{table}

To understand why naive imitation learning fails on gigapixel WSI navigation, we conduct two diagnostic studies dissecting (i) within-episode visual evidence integration (Table~\ref{tab:zoom_diagnostics}) and (ii) trajectory distillation shortcuts (Table~\ref{tab:shortcut_ablation}).

\subsubsection{Breaking the Step-0 Overview Shortcut (Table~\ref{tab:zoom_diagnostics})}
In Table~\ref{tab:zoom_diagnostics}, evaluating the baseline static-option bilinear head (M0, which pairs the 28-layer terminal hidden query $h_t$ with context-free Layer-0 option embeddings) before zooming (Step 0) versus after zooming (Step $T$) exposes a striking rigidity: M0 produces the exact same prediction at Step $T$ as at Step 0 on $98.58\%$ of all $351$ test episodes ($346/351$), with a negligible $L_1$ probability shift of $0.021$ ($\Delta_{\text{zoom}} = +2$). Despite executing $3.12$ crop reads per query, M0 relies almost entirely on the Step-0 overview thumbnail $I_{\text{ov}}$ rather than revising its belief from newly acquired crops.

Augmenting the static bilinear keys with our Gated 28-Layer Contextual-Bilinear Answer Head (M1, blending $z_{\text{bil}}$ with mean-centered contextual LM letter logits $z_{\text{ctx}}$) drops Step-0/Step-$T$ agreement from $98.58\%$ to $64.39\%$ and lifts $L_1$ belief shift $18\times$ to $0.375$, yielding $+1.42\text{ pp}$ over M0 terminal accuracy ($47.01\% \to 48.43\%$) and a $+2.28\text{ pp}$ Step-0 to Step-$T$ gain ($46.15\% \to 48.43\%$, $\Delta_{\text{zoom}} = +8$). Adding explicit magnification tags and tail re-anchoring (M2) and overview dropout (M3) reveals a non-monotonic interaction: M2 and M3 break thumbnail rigidity (down to $58.69\%$ agreement and $0.424$ $L_1$ shift, with M2 reaching $64.04\%$ on Bucket D), yet still yield negative net zoom flips ($-8$ and $-6$) on their own because unaligned trajectories induce noisy late revisions. Only when paired with counterfactual trajectory alignment (JevNavAgent v5) does active zooming unlock a $+35$ net correction ($+74$ questions corrected vs.\ $-39$ lost), lifting accuracy from $39.89\%$ at Step 0 to $49.86\%$ at Step $T$ ($+9.97\text{ pp}$). Stage-2 Traj-GRPO further raises Step-0 to $42.74\%$ and Step-$T$ to $50.43\%$ ($\Delta_{\text{zoom}} = +27$, $+7.69\text{ pp}$).

\subsubsection{Ablating Trajectory Distillation Shortcuts on CAMELYON16 (Table~\ref{tab:shortcut_ablation})}
Table~\ref{tab:shortcut_ablation} compares four training regimes across all $46$ CAMELYON16 Dev slides ($23$ Pos + $23$ Neg):
\begin{enumerate}
    \item Positive-Only Distillation (Row 1): Training solely on positive target-seeking rollouts ($1{,}042\text{:}0$ CAM stops, $66\text{:}0$ lateral pan stops) teaches direct coarse-to-fine lesion search ($21.74\%$ positive localization hit rate across $3.57/3.00$ reads) but zero negative exclusion capability ($0.00\%$ specificity, $50.00\%$ accuracy), as the policy predicts positive metastasis on every normal slide.
    \item Unaligned VLM Trajectory Distillation (Row 2): Directly adding raw negative VLM rollouts leaves lateral panning sequences $(16 \to 4 \to 4)$ exclusively in negative trajectories ($0\text{ Pos} : 70\text{ Neg}$). The student exploits short horizon length ($t \le 4$) as a positive trigger ($3.83/8.17$ reads), degrading positive localization to $13.04\%$ while raising specificity only modestly from $0.00\%$ to $30.43\%$ ($65.22\%$ slide accuracy).
    \item Fresh-Initialized Intermediate READ Negative Supervision (Row 3): Enforcing multi-region exclusion ($\mathcal{M}_{\text{neg}}$) and kinematic alignment ($12\text{:}13$ lateral pan stops) in a fresh-initialized run ($1{,}200$ steps, $\text{lr}=5\times 10^{-5}$) fixes trajectory exploration ($11.78/12.00$ reads), but supervising the answer head with ``Negative'' targets on intermediate $\text{READ}$ steps introduces an $87.5\% : 12.5\%$ ($7:1$) READ-to-STOP state predominance ($2{,}450$ READ vs.\ $349$ STOP records) and collapses positive sensitivity to $0.00\%$ ($100.0\%$ specificity).
    \item M3-Warm-Started Counterfactual Alignment with Stop-Decoupled Supervision (Row 4): Warm-starting from M3 ($600$ steps, $\text{lr}=2\times 10^{-5}$) and supervising the answer head strictly on $\text{STOP}$ states simultaneously achieves $30.43\%$ positive localization hits ($7/23$), $95.65\%$ negative specificity ($22/23$), and $71.74\%$ overall slide accuracy ($33/46$).
\end{enumerate}

\begin{figure*}[t]
\centering
\includegraphics[width=0.98\textwidth]{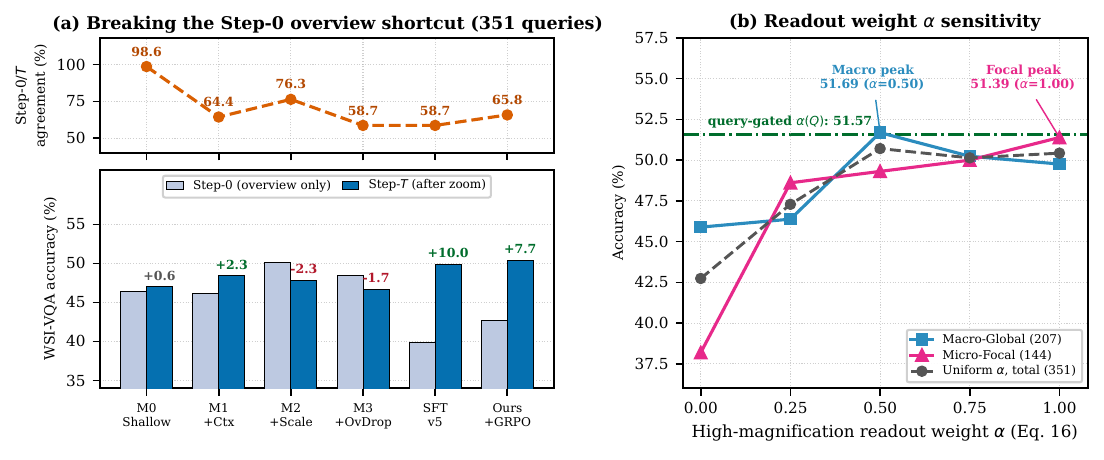}
\caption{Within-episode zoom shortcut diagnostics and multi-scale readout analysis on Clean Visual WSI-VQA ($351$ queries). \textbf{(a)} Progressive architectural and counterfactual debiasing (M0 $\to$ M1 $\to$ M2 $\to$ M3 $\to$ SFT v5 $\to$ +Traj-GRPO) reduces Step-0/Step-$T$ prediction rigidity from $98.58\%$ to $58.69\%\text{--}65.81\%$ and unlocks a $+7.69$ to $+9.97$ pp Step-$T$ accuracy gain from active zooming ($39.89\% \to 49.86\%$ on SFT v5 and $42.74\% \to 50.43\%$ on +Traj-GRPO). \textbf{(b)} Varying the high-magnification readout weight $\alpha \in [0, 1]$ on our Stage-2 policy reveals an inverted-U curve on Macro-Global queries (peaking at $\alpha=0.50$, $51.69\%$) versus a monotonic climb on Micro-Focal cellular queries (peaking at $\alpha=1.00$, $51.39\%$); our query-gated readout $\alpha_{m_Q}$ (Eq.~\eqref{eq:multiscale_readout}) captures both optima ($51.57\%$ overall).}
\label{fig:zoom_multiscale}
\end{figure*}

\begin{table}[t]
\centering
\footnotesize
\caption{Ablation of Stage-2 Trajectory-Level GRPO Checkpoints Across WSI-VQA ($351$) and CAMELYON16 Official Test ($129$, $B=12$) ($\%$). Each row evaluates a single saved checkpoint.}
\label{tab:grpo_ablation}
\renewcommand{\arraystretch}{1.16}
\setlength{\tabcolsep}{0.95pt}
\begin{tabular}{l>{\columncolor{priCol}}cc>{\columncolor{priCol}}cc>{\columncolor{priCol}}c}
\toprule
\textbf{Stage-2 RL Checkpoint} & \textbf{VQA$^\star$} & \textbf{Focal} & \textbf{Loc$^\star$} & \textbf{Spec} & \textbf{Acc$^\star$} \\
\midrule
1.\ Stage-1 SFT (v5)   & $49.86$ & $49.31$ & $10.20$ & $88.75$ & $66.67$ \\
2.\ + GRPO (Heads)     & $48.43$ & $49.31$ & $0.00$  & $\mathbf{100.0}$ & $62.02$ \\
3.\ + GRPO (WiSE-FT)   & $47.29$ & $47.92$ & $20.41$ & $88.75$ & $68.99$ \\
\rowcolor{oursRow}
4.\ \textbf{+ GRPO (CAM16)} & $46.15$ & $44.44$ & $\mathbf{22.45}$ \gain{12.3} & $93.75$ & $\mathbf{69.77}$ \gain{3.1} \\
\rowcolor{oursRow}
5.\ \textbf{+ GRPO (VQA+KL)} & $\mathbf{51.57}$ \gain{1.7} & $\mathbf{51.39}$ & $14.29$ \gain{4.1} & $82.50$ & $61.24$ \\
\bottomrule
\end{tabular}\\[2pt]
{\scriptsize Note: Rows 1--4 report pure Step-$T$ VQA accuracy; Row 5 reports query-gated multi-scale VQA ($51.57\%$; pure Step-$T$ is $50.43\%$). Row 2 freezes the entire 2B backbone; Row 4 specializes in polygon-guided stopping on CAM16; Row 5 optimizes $\Delta P_t$ with soft KL lock on WSI-VQA, illustrating the task-specialization trade-off between screening and open-ended VQA.}
\end{table}

\subsubsection{Ablating Stage-2 Trajectory-Level GRPO Checkpoints (Table~\ref{tab:grpo_ablation})}
Table~\ref{tab:grpo_ablation} evaluates each saved Stage-2 reinforcement learning checkpoint across both benchmarks:
\begin{enumerate}
    \item Necessity of Language-Layer LoRA Backbone Adaptation (Row 2): Freezing the 2B vision-language backbone and updating only the factorized decision heads (+ GRPO (Heads)) fails on localization ($0.00\%$ Loc Hit, $100.0\%$ Spec): because benign lymphoid follicles outnumber tumor patches along exploration rollouts, a frozen feature space forces the decision heads to shift their global bias toward ``Negative.'' Jointly updating the language-layer LoRA adapters (Rows 3--5) enables visual separation between malignant clusters and reactive germinal centers.
    \item Polygon-Guided Branch/Stop Credit vs.\ Domain Specialization (Rows 3 and 4): Optimizing full-LoRA trajectories with branch search reward and immediate in-polygon stopping on CAMELYON16 (Row 4, + GRPO (CAM16)) lifts Test localization hit rate to $22.45\%$ ($91.7\%$ stop conversion, vs.\ $10.20\%$ SFT Online and $14.29\%$ SFT Calib.) and raises slide accuracy to $69.77\%$, while WiSE-FT weight-space interpolation~\cite{wortsman2022robust} (Row 3) balances both tasks at $20.41\%$ Test localization and $68.99\%$ slide accuracy.
    \item Step-Wise Confidence Gain $\Delta P_t$ and Soft KL Lock (Row 5): Optimizing step-wise confidence gain $\Delta P_t$ with soft KL distillation locking (+ GRPO (VQA+KL)) preserves four-way option discrimination while lifting Micro-Focal accuracy to $51.39\%$. On overall WSI-VQA accuracy, online RL contributes $+0.57\text{ pp}$ under pure Step-$T$ readout ($49.86\% \to 50.43\%$), and query-gated multi-scale readout adds a further $+1.14\text{ pp}$ to reach $51.57\%$ (while lifting CAMELYON16 Test localization from $10.20\%$ to $14.29\%$ at $82.50\%$ specificity).
\end{enumerate}

\subsection{Qualitative Multi-Scale Navigation Analysis}
\label{subsec:exp_qualitative}

\begin{figure*}[t]
\centering
\includegraphics[width=0.96\textwidth]{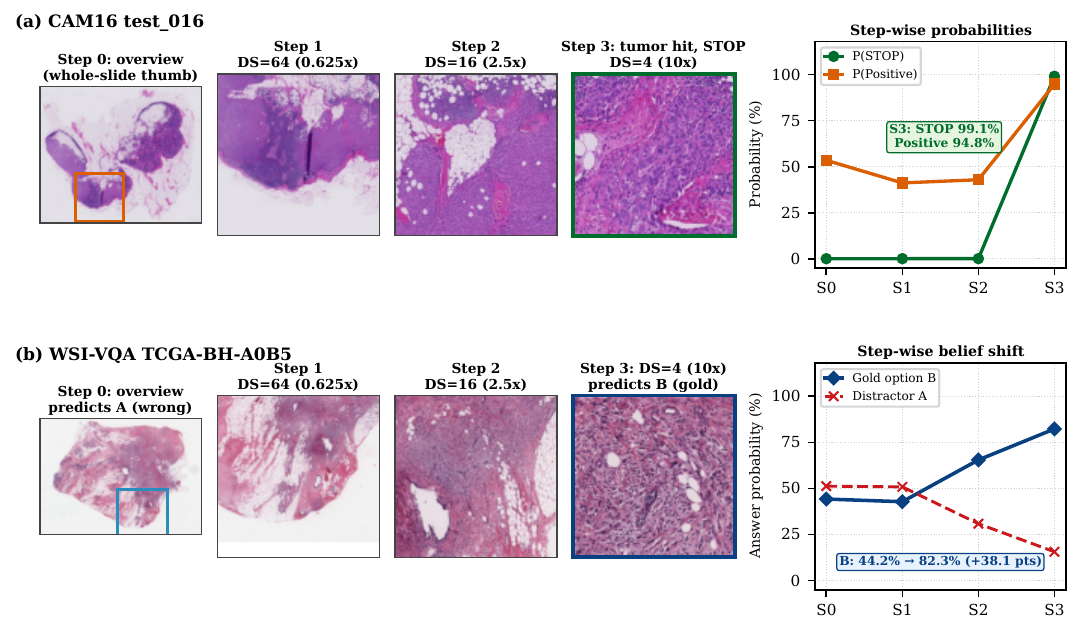}
\caption{Qualitative multi-scale navigation trajectories and step-by-step probability progression from held-out evaluation episodes. \textbf{(a)} On CAMELYON16 Official Test case 016, JevNavAgent + Traj-GRPO (CAM16) initiates from the resized whole-slide overview, drills down through $\text{DS}=64 \to 16 \to 4$ into the metastasis locus, and immediately triggers $\text{STOP}(\text{Positive})$ at Step 3 as $P(\text{STOP})$ jumps from $0.1\%$ to $99.1\%$ and $P(\text{Positive})$ reaches $94.8\%$. \textbf{(b)} On Clean Visual WSI-VQA Micro-Focal query on TCGA-BH-A0B5 (``Is there any lymphovascular invasion identified?''; Option A: Yes, Option B: No), JevNavAgent + Traj-GRPO (VQA+KL) initially mispredicts Option A ($51.2\%$ vs.\ $44.2\%$) from the Step-0 whole-slide overview; active zooming ($\text{DS}=64 \to 16 \to 4$, $10\times$) overrides this affirmative prior with a $+38.1\text{ pp}$ confidence surge on the ground-truth Option B ($44.2\% \to 82.3\%$).}
\label{fig:qualitative_traj}
\end{figure*}

Fig.~\ref{fig:qualitative_traj} illustrates how Stage-2 Traj-GRPO translates multi-scale visual evidence into grounded decisions across both tasks: on CAMELYON16 (Fig.~\ref{fig:qualitative_traj}a), the policy narrows from the resized overview and initial $\text{DS}=64$ crop to subcapsular cellular resolution ($\text{DS}=4$, $10\times$) and halts immediately inside the tumor boundary rather than stopping prematurely in benign follicles; on focal VQA (Fig.~\ref{fig:qualitative_traj}b), the Gated Contextual-Bilinear Answer Head actively revises its diagnostic belief upon inspecting $10\times$ stromal vessels, directly countering low-resolution affirmative hallucination bias.

\section{Conclusion}
\label{sec:conclusion}
We presented JevNavAgent, a 2B-parameter single-forward visual navigation agent for gigapixel whole-slide pathology. Grounded in the Jev (System One) paradigm, JevNavAgent outputs factorized stopping, spatial-candidate, and gated contextual-bilinear answer distributions in one forward pass per state ($0$ generated tokens), delegating candidate generation, parent re-entry, and root switching to deterministic code. Eliminating autoregressive decoding enables both fast inference ($5.63\,\text{s}$ per episode, $11.8\times\text{--}12.4\times$ faster than frontier VLM agents) and single-GPU online trajectory-level GRPO ($1.0\text{--}4.2\,\text{h}$) atop offline counterfactual trajectory alignment. On Clean Visual WSI-VQA ($351$ queries), JevNavAgent reaches $50.43\%$ Step-$T$ ($51.57\%$ multi-scale) overall accuracy and $51.39\%$ focal accuracy. On CAMELYON16 ($311$ WSIs), Stage-1 SFT attains $30.43\%$ Dev localization with zero-shot budget scaling; on Official Test, Stage-2 Traj-GRPO Online achieves the highest Strict Loc+Spec ($67.44\%$) and Calib.\ achieves the highest localization ($22.45\%$, $50.00\%$ on macrometastases) and slide accuracy ($69.77\%$), outperforming SFT under both matched stopping rules. Current limitations include breast-only cohorts (TCGA-BRCA and CAMELYON16), two task-specialized Stage-2 checkpoints, and micrometastasis localization under a $12$-read budget. Future work will pursue unified multi-task RL, higher-recall search, and multi-stain, multi-organ generalization.

\bibliographystyle{IEEEtran}
\bibliography{references}

\end{document}